\documentclass{article}

\usepackage{microtype}
\usepackage{graphicx}
\usepackage{subfigure}
\usepackage{booktabs} 
\usepackage[table]{xcolor}
\usepackage{pifont}

\usepackage{hyperref}

\usepackage[accepted]{icml2023}
\usepackage{amsmath}
\usepackage{amssymb}
\usepackage{mathtools}
\usepackage{amsthm}
\usepackage{bm}
\usepackage{makecell}

\usepackage[capitalize,noabbrev]{cleveref}

\theoremstyle{plain}

\theoremstyle{definition}

\theoremstyle{remark}

\usepackage[textsize=tiny]{todonotes}

\newcommand{\textpara}[2]{\noindent\textbf{#1}. {#2}\vspace{0.1in}}

\newcommand{\tb}[1]{\textbf{#1}}
\newcommand{\ccost}{\mathcal{C}}
\newcommand{\video}{\mathcal{V}}
\newcommand{\videob}{\bar{\mathcal{V}}}
\newcommand{\query}{\mathcal{Q}}
\newcommand{\response}{\mathcal{R}}
\newcommand{\lossem}{\mathcal{L}_\textrm{EM}}
\newcommand{\losssel}{\mathcal{L}_\textrm{SEL}}
\newcommand{\coefsel}{\lambda_\textrm{SEL}}
\newcommand{\lossfd}{\mathcal{L}_\textrm{FD}}
\newcommand{\coeffd}{\lambda_\textrm{FD}}
\newcommand{\losspd}{\mathcal{L}_\textrm{PD}}
\newcommand{\coefpd}{\lambda_\textrm{PD}}
\newcommand{\losstot}{\mathcal{L}_\textrm{total}}
\newcommand{\vfeat}{\bm{v}}
\newcommand{\vfeatb}{\bar{\bm{v}}}
\newcommand{\sfeat}{\bm{s}}
\newcommand{\qfeat}{\bm{q}}
\newcommand{\cfeat}{\bm{c}}
\newcommand{\cfeatb}{\bar{\bm{c}}}
\newcommand{\svfeat}{\sfeat \oplus \vfeatb}
\newcommand{\bfeatb}{\bar{\bm{b}}}
\newcommand{\yes}{\ding{51}}

\newcommand{\tbr}[1]{\textbf{\color{red}{#1}}}

\newcolumntype{P}[1]{>{\centering\arraybackslash}p{#1}}

\newcommand{\tacos}{TACoS}

\icmltitlerunning{SpotEM: Efficient Video Search for Episodic Memory}

\newcommand{\modelname}{SpotEM} 

\begin{document}

\twocolumn[
\icmltitle{\modelname: Efficient Video Search for Episodic Memory}



\icmlsetsymbol{equal}{*}

\begin{icmlauthorlist}
\icmlauthor{Santhosh Kumar Ramakrishnan}{uta}
\icmlauthor{Ziad Al-Halah}{uofu}
\icmlauthor{Kristen Grauman}{uta,mai}
\end{icmlauthorlist}

\icmlaffiliation{uta}{UT Austin}
\icmlaffiliation{mai}{FAIR, Meta AI}
\icmlaffiliation{uofu}{University of Utah}

\icmlcorrespondingauthor{S. Ramakrishnan}{sramakrishnan@utexas.edu}

\icmlkeywords{Episodic Memory, Egocentric Perception, Ego4D}

\vskip 0.3in
]


\printAffiliationsAndNotice{}  

\begin{abstract}
The goal in episodic memory (EM) is to search a long egocentric video to answer a natural language query (e.g., \emph{``where did I leave my purse?"}). Existing EM methods exhaustively extract expensive fixed-length clip features to look everywhere in the video for the answer, which is infeasible for long wearable-camera videos that span hours or even days. We propose \modelname{}, an approach to achieve efficiency for a given EM method while maintaining good accuracy.  \modelname{} consists of three key ideas: 1) a novel clip selector that learns to identify promising video regions to search conditioned on the language query; 2) a set of low-cost semantic indexing features that capture the context of rooms, objects, and interactions that suggest where to look; and 3) distillation losses that address the optimization issues arising from end-to-end joint training of the clip selector and EM model. Our experiments on 200+ hours of video from the Ego4D EM Natural Language Queries benchmark and three different EM models demonstrate the effectiveness of our approach: computing only 10\% -- 25\% of the clip features, we preserve 84\% -- 97\% of the original EM model's accuracy. Project page: {\small \url{https://vision.cs.utexas.edu/projects/spotem}}

\end{abstract}

\section{Introduction}
\label{sec:intro}

The limitations of human memory can pose an obstacle for our day-to-day activities. We forget where we put things (\emph{``where did I leave my car keys?"}), fail to notice the state of particular objects (\emph{``did I turn off the stove? how much milk is left in the fridge?"}), and struggle to recall details about past events (\emph{``who did I run into while jogging last week? what was the name of the bakery where we bought the muffins?}"). First-person or ``egocentric" perception from wearable devices like augmented reality (AR) glasses could remove that cognitive load, allowing a user to ask questions on-the-fly about their own past visual experience.  Similarly, an intelligent mobile robot could relay valuable information about what it has seen based on people's conversational queries (\emph{``did anyone feed the dog yet?"}).

This vision of a \emph{personal episodic memory} prompts interesting new challenges in computer vision and multimodal learning. Not only does it require recognizing objects and activities from the first-person perspective, but also identifying which visual content is sufficient to answer a question expressed in free-form natural language, accounting for the fact that the correct answer may occupy only a tiny portion of the entire video.  Moreover, all this must be done in a scalable manner, given that the visual history of a user will ultimately span hours, days, weeks, or more.

\begin{figure}[t]
    \centering
    \includegraphics[width=0.48\textwidth,clip,trim={0 6.0cm 9.7cm 0}]{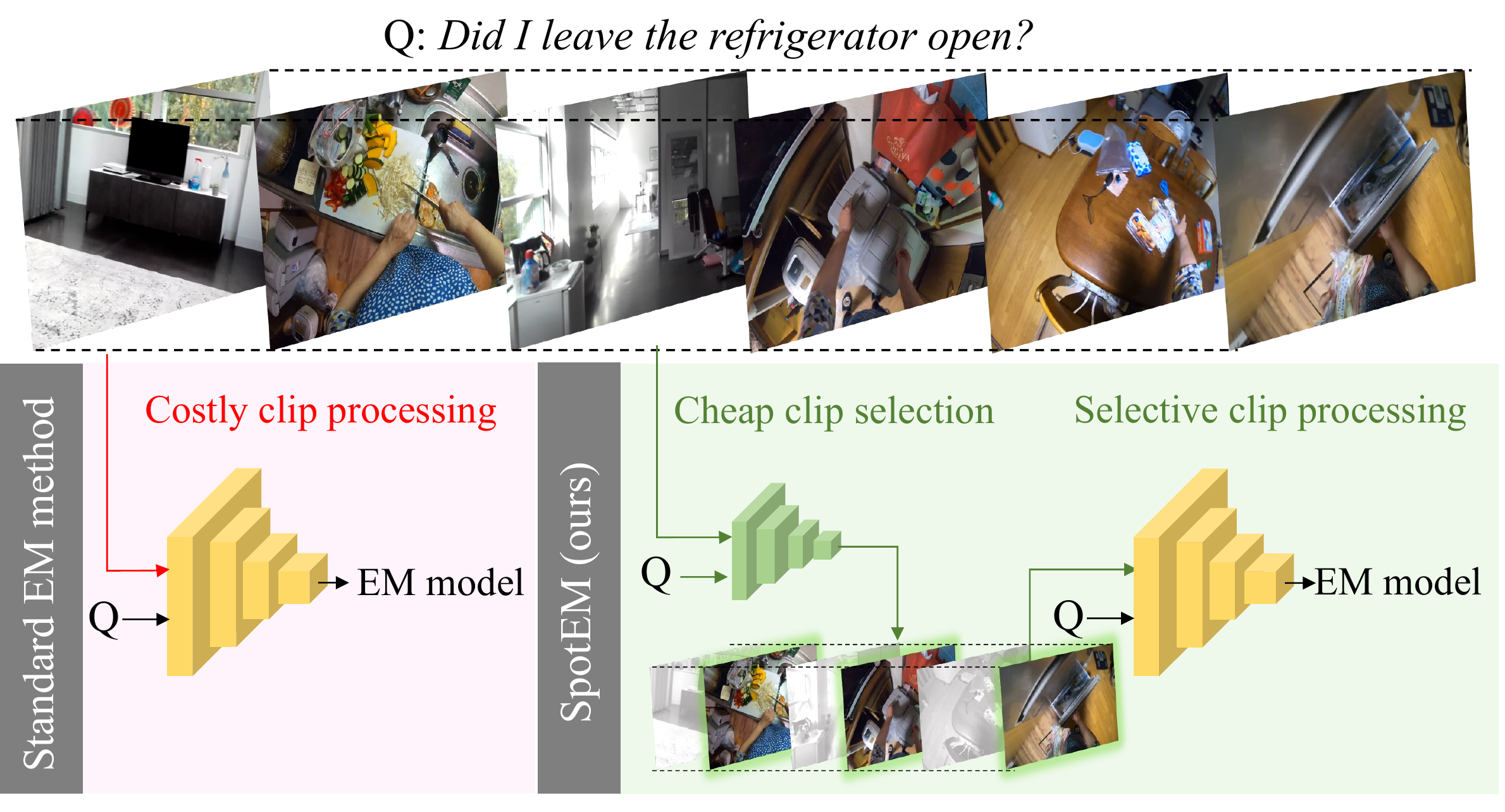}
    \vspace*{-0.25in}
    \caption{Standard EM methods (see left) divide videos into fixed-length clips and perform costly processing of every clip. The processed clips are provided to the EM model for query-conditioned search. We propose \modelname, a clip-selection approach that spots clips relevant to the query cheaply, and selectively processes these clips to serve as inputs to the EM model (see right).
    } 
    \vspace*{-0.15in}
    \label{fig:intro}
\end{figure}

The recently introduced Episodic Memory (EM) benchmark\footnote{In particular, we focus on Ego4D's NLQ, the EM variant with natural language queries, as opposed to other variants defined with object or activity queries.} from Ego4D targets this task~\cite{grauman2022ego4d}: given a natural language query and a long egocentric video, identify the precise temporal window containing the answer. The EM benchmark has attracted significant attention and is the subject of community-wide competitions~\cite{grauman2022ego4d,liu2022reler,lin2022egocentric,hou2022efficient,mo2022simple,chen2022internvideo} attracting dozens of submissions in the first year alone. Hallmarks of the benchmark are the free-form nature of the natural language queries, the long-form nature of the egocentric videos (8.2 minutes on average), and the short responses that span a tiny fraction of the overall video (2\% on average). The result  is a ``needle-in-the-haystack" problem for video localization. Importantly, these facets set the EM task apart from traditional activity recognition and language grounding tasks, which feature much shorter videos in which most content is relevant~\cite{caba2015activitynet,FCVID,gao2017tall} (e.g., 30 second video in which a third of it displays the target action). 

While current EM methods have made exciting headway, they neglect the practical scaling issue that is so central to episodic memory. Today's methods extract expensive spatio-temporal features for densely sampled clips throughout the video~\cite{feichtenhofer2019slowfast,bertasius2021space,tong2022videomae,girdhar2022omnivore}. This step alone contributes over 99.9\% of an EM model's computational cost.\footnote{A single clip feature from ~\citet{chen2022internvideo} consumes 2.09 TFLOPs, while EM modules from ~\citet{zhang2020span} (excluding the video backbone) consume 3 GFLOPs for the whole video.} Such an approach becomes intractable as the video length grows, especially for real-time applications like AR and robotics where the constrained on-board computation severely limits the ability to operate such heavy-weight models. Besides, EM is just one of several functions that need to be supported by these devices---all resources cannot be devoted to just computing features for EM. 

We observe that 1) not all parts of the video are useful for reasoning about a given query, and 2) there are high-level visual semantics about rooms, objects and interactions that could steer our attention towards where to look. For example, given the queries, \emph{``did I leave the lights on in the living room?"} or \emph{``did I close the refrigerator?"}, we can ignore video clips recorded in rooms other than the living room or kitchen, respectively. Notably, these associations cannot be neatly enumerated, however, given the free-form nature of the queries. In the query \emph{``who was I with when I first boarded the golf cart?"}, the chain of reasoning becomes much more complex since we need to reason about all instances of boarding a golf cart, and identify those instances where there was someone else with you. This points to the need for learning query-conditioned priors that can use such high-level semantics to help narrow down the search task.

We build upon these intuitions to propose \modelname{}, a novel approach to make a given EM method efficient. See~\cref{fig:intro}. The idea is to preview the video using cheap indexing features, intelligently select a small subset of \emph{query-relevant} clips, and only use these clips for the full EM search. This can cut down computational costs without sacrificing model performance. Our approach differs from prior work for efficient action recognition~\cite{wu2018compressed,yeung2016end,wu2019liteeval,feichtenhofer2020x3d,gao2020listen} and video captioning~\cite{chen2018less,suin2020efficient} that assumes much shorter inputs and does not condition on free-form language queries. 

To tackle this challenging setting with long-form egocentric videos and language queries, we design \emph{MemorySpotter}, a novel clip selection architecture that uses a cross-modal transformer to recursively preview the video and identify query-relevant clips. We further design a set of inexpensive semantic-indexing features that capture video context about \textbf{r}ooms/scenes visited, \textbf{i}nteractions observed, and the \textbf{o}bjects present (\emph{RIO}), thereby exposing to the model the high-level visual context that may support the free-form query text. Further, we propose expert-based distillation losses to address optimization issues arising from jointly training MemorySpotter and the EM modules. Our experiments on the Ego4D EM benchmark demonstrate that our approach is effective and versatile: when tested with multiple EM methods, it achieves 95+\% of the original EM method's performance while computing heavy-weight video features for only 25\% of the clips. We also perform ablation studies to validate the design of \modelname{} and present a detailed study of its clip selection behaviors.

\section{Related work}

\textpara{Egocentric video understanding}{
Unlike internet-style data, egocentric video captures the camera wearer's perspective of their activities, and is the subject of multiple datasets~\cite{fathi2011gtea,kazakos2019epic,furnari2020rolling,Damen2022RESCALING,grauman2022ego4d}. Ego-video raises new research problems in human-object interaction~\cite{cai2018understanding,damen2014you}, activity recognition~\cite{kazakos2019epic,zhou2015temporal}, anticipation~\cite{abu2018will,girdhar2021anticipative}, video summarization~\cite{del2016summarization,yongjae-ijcv2015}, and spatial organization~\cite{ortis2017organizing,furnari2016recognizing,nagarajan2020ego,price2022unweavenet}. We focus on the recently introduced episodic memory task, which requires answering queries about long-form egocentric videos~\cite{grauman2022ego4d}.} 

\textpara{Episodic memory}{
The goal in episodic memory is to temporally localize the response to natural language queries. This task extends video question answering~\cite{xu2017video,rohrbach2017movie,xu2021vlm,zhang2020span} to the challenging egocentric setting. The EM benchmark offers a compelling step towards episodic memory applications in AR and has been the subject of multiple challenges at top conferences~\cite{grauman2022ego4d}.\footnote{\scriptsize CVPR 2022: \url{https://ego4d-data.org/workshops/cvpr22/}\\ ECCV 2022: \url{https://ego4d-data.org/workshops/eccv22/}} Prior work adapts video-language grounding methods such as 2D-TAN~\cite{zhang2020learning}, VSLNet~\cite{zhang2020span}, and VSLNet-L~\cite{zhang2021natural} to perform EM search, and the state-of-the-art methods develop video representations~\cite{lin2022egocentric,chen2022internvideo} and data augmentation strategies~\cite{liu2022reler}. Unlike prior work which is purely motivated by accuracy, we tackle the orthogonal problem of search efficiency, which is critical for real-world EM applications. Our results show \modelname{}'s versatility: deployed to augment three popular EM models~\cite{chen2022internvideo,liu2022reler,lin2022egocentric}, it successfully achieves substantial efficiency gains for each one.
}

\textpara{Efficient video models}{
Prior work develops efficient video recognition architectures through compute-efficient modules~\cite{feichtenhofer2020x3d}, feature-distillation~\cite{zhang2016real,gao2020listen}, compressed video processing~\cite{zhang2016real,wu2018compressed}, or adaptively choosing between cheap and expensive inputs and modules~\cite{zhu2020faster,meng2020ar,li20212d}. Orthogonal to this are frame-sampling methods that select informative subsets of the video for video recognition ~\cite{chen2011dynamic,yeung2016end,wu2019liteeval}, typically by first previewing the video with an inexpensive module. For example, \citet{chen2011dynamic} use an inexpensive background subtraction module to filter out irrelevant frames, while \citet{korbar2019scsampler} predict each clip's saliency and select the most salient clips for processing. Alternatively, reinforcement learning (RL) can be used to learn frame-selection policies for action recognition and detection \cite{yeung2016end,fan2018watching}. \citet{wu2019liteeval} improve over RL methods by using the Gumbel-Softmax trick to reduce frame selection to a supervised learning problem. Some methods further leverage audio to guide the sampling process~\cite{jiang2015super,gao2020listen,panda2021adamml}. 

Rather than sequentially processing one clip at a time~\cite{wu2019adaframe,yeung2016end,wu2019liteeval}, our clip selection architecture simultaneously previews all of the video at once using cheap image features and makes selection choices among all clips. Furthermore, prior models that do parallel sampling preview the entire video only once to make sampling decisions~\cite{korbar2019scsampler,lin2022ocsampler}. We instead propose a \emph{recursive} approach that previews the video multiple times to actively grow the set of observed clips. At each recursive step, we select a subset of clips, extract their (heavier) features, and incorporate this knowledge for future clip selection steps. We demonstrate the significance of these contributions in our experiments. Beyond architectural improvements over prior work, \modelname{} also has novel contributions of designing a) semantic index features relevant to EM and b) distillation losses to address optimization limitations arising from jointly training a clip selector and the task models.
}
\section{Approach}

We propose \modelname{}, a clip-selection approach that intelligently spots query-relevant clips for efficient EM. Our overall approach consists of a novel clip selection architecture called MemorySpotter,  our RIO semantic indexing features to select clips relevant for EM, and distillation losses to address optimization issues arising from jointly training MemorySpotter with EM task modules. MemorySpotter previews the video using our RIO features, which are obtained by selecting a single image from each clip and encoding them using efficient image encoders~\cite{tan2019efficientnet}. Heavy clip features are then extracted from only the smaller subset of clips selected by MemorySpotter. Next, we review the EM task definition and discuss our approach. 

\subsection{Episodic memory task}
\label{sec:em_task}

The goal in natural-language episodic memory is to perform query-driven reasoning about long-form egocentric videos~\cite{grauman2022ego4d}. Formally, given an egocentric video $\video$ capturing a camera wearer's past experiences and a query text $\query$, the task requires temporally localizing where the answer can be seen in the video, i.e., a response window $\response$ = $[t_s, t_e]$ defined by start $t_s$ and end $t_e$. 

Recent work adapts video-language grounding models like VSLNet~\cite{zhang2020span} and VSLNet-L~\cite{zhang2021natural} to achieve state-of-the-art results for EM~\cite{lin2022egocentric,liu2022reler,chen2022internvideo}. Our analysis shows that these methods devote over 99.9\% of their computation to extracting spatio-temporal features for fixed-length clips that are sampled densely from the video. We instead propose to preview the video cheaply and identify query-relevant clips to perform EM efficiently.

\begin{figure*}[t]
    \centering
    \includegraphics[width=\textwidth,clip,trim={0 6.5cm 6cm 0}]{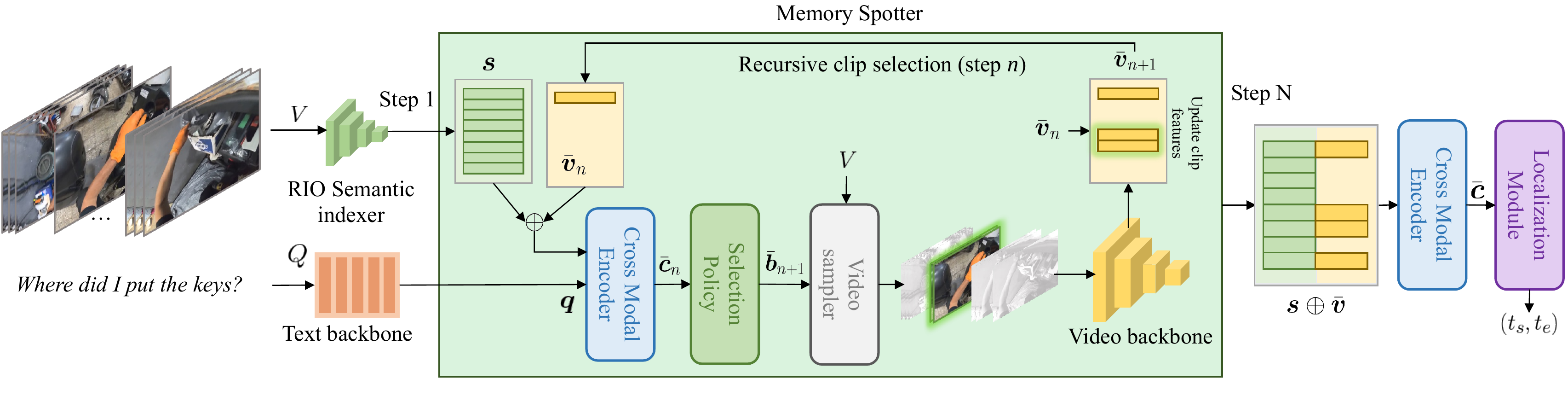}
    \vspace*{-0.25in}
    \caption{\textbf{SpotEM for efficient video search in episodic memory}: We enhance standard EM models with the ability to intelligently select clips for expensive feature extraction, with the aim of reducing computational cost without affecting accuracy. First, we extract our RIO semantic indexing features $\sfeat$ that capture room, interaction, and object context in the video using a cheap semantic indexer. We then encode the query using a text backbone to obtain query features $\qfeat$ (see left). The MemorySpotter module operates recursively to select a subset of clips relevant to the query (see center). It first previews the video using semantic index $\sfeat$. It then alternates between selecting subsets of video clips $\videob_n$ for expensive feature extraction, and previewing the video again with both the semantic index $s$ and the previously selected clip features $\vfeatb_{n+1}$. After $N$ steps of the MemorySpotter, we use its cumulative selected clip features $\vfeatb = \vfeatb_{N+1}$ and semantic index $\sfeat$ to perform EM (see right). A cross-modal encoder jointly reasons about the concatenated features $\svfeat$ and $\qfeat$, to obtain a cross-modal embedding $\cfeatb$. A localization module then predicts the temporal extents of the response. Note that the cross-modal encoder inside MemorySpotter shares weights with the cross-modal encoder on the right. At step $n = 1$, no clips are selected, i.e., $\bar{\bm{v}}_1 = \vec{\bm{0}}$.
    }
    \label{fig:nlq_arch}
\end{figure*}

\subsection{SpotEM for efficient episodic memory}
\label{sec:eem_desc}

We now provide an overview of our SpotEM approach. See~\cref{fig:nlq_arch}. The model inputs consist of the video $\video$ with clips $[\video_1, \cdots, \video_L]$, and a text query $\query$ with words $[\query_1, \cdots, \query_T]$. First, a pretrained semantic indexer, which consists of one or more image encoders, is used to extract semantic index features $\sfeat \in \mathbb{R}^{L \times D_s}$.
\begin{equation}
    \label{eqn:indexbackbone}
    \sfeat = [s_1, s_2, \cdots, s_L] = \textrm{SemanticIndexer}(\video).
\end{equation}
These are image features extracted by sampling one image within each video clip, and they are inexpensive to compute (2-3 orders lower cost than clip features). These will serve as an initial preview of the video for intelligent clip selection (more details in~\cref{sec:rio_feats}).

A pretrained DistillBERT backbone~\cite{sanh2019distilbert} is then used to extract the query text features $\qfeat \in \mathbb{R}^{T \times D_q}$.
\begin{equation}
    \label{eqn:textbackbone}
    \qfeat = [q_1, q_2, \cdots, q_T] = \textrm{TextBackbone}(Q).
\end{equation}
The MemorySpotter module uses the semantic index $\sfeat$, query features $\qfeat$, and the video $\video$ to recursively identify a final subset of video clips $\videob$ that are relevant to the query. The expensive clip features for $\videob$ are obtained using pretrained video backbones like Timesformer~\cite{bertasius2021space} and VideoMAE~\cite{tong2022videomae} (see Sec.~\ref{sec:exp_setup}).\footnote{The video features for clips not selected are set to zeros in $\vfeatb$.}
\begin{equation}
    \label{eqn:videobackbone}
    \vfeatb = \textrm{VideoBackbone}(\videob) \in \mathbb{R}^{L \times D_v}
\end{equation}

A cross-modal encoder uses the concatenated clip and semantic index features $\svfeat$ and the query features $\qfeat$ to perform cross-modal reasoning to enhance the video features with query-specific information. 
\begin{equation}
    \label{eqn:crossmodal}
    \cfeatb = \textrm{CrossModalEncoder}(\svfeat, \qfeat) \in \mathbb{R}^{L \times D_h}
\end{equation}
Finally, a localization module is used to predict the temporal extent of the response $\hat{\mathcal{R}}$ based on  $\cfeatb$: 
\begin{equation}
    \label{eqn:localize}
    \hat{\mathcal{R}} = [\hat{t}_s, \hat{t}_e] = \textrm{LocalizationModule}(\cfeatb).
\end{equation}
This formulation augments existing EM methods to be efficient. Specifically, EM methods like VSLNet~\cite{zhang2020span} and VSLNet-L~\cite{zhang2021natural} use video backbones to extract clip features, a text backbone to extract query features, a cross-model encoder to jointly encode the two modalities, and a localization module to predict the response locations (see~\cref{suppfig:em_arch}). SpotEM modulates the video inputs to these EM models by intelligently selecting a subset of clips for heavy clip feature extraction. While the architectural details of the cross-modal encoder and localization modules vary across EM methods (described in~\cref{suppsec:em_task_archs}), our approach remains unchanged.

SpotEM achieves efficiency by previewing the video cheaply using the semantic indexer (detailed in Sec.~\ref{sec:rio_feats}) and then recursively selecting a subset of clips relevant to the query using MemorySpotter, which we describe next.

\subsection{MemorySpotter architecture description}
\label{sec:selection_desc}
One of our key contributions is the MemorySpotter architecture for intelligent clip selection for EM. See~\cref{fig:nlq_arch}~(center). It first previews the entire video using low-cost semantic index ($\sfeat$). It then alternates between selecting video clips for expensive feature extraction, and previewing the video again with both the semantic indexing features \emph{and} the previously selected clip features ($\vfeatb_{n}$).

Specifically, let $n \in [1, \cdots, N]$ denote the recursive step and $\vfeatb_n \in \mathbb{R}^{L \times D_v}$ denote the clip features for the subset of video clips selected from steps $1$ to $n-1$, where $\vfeatb_1~=~[0]_{L \times D_v}$ is a matrix of zeros. The semantic indexing features $\sfeat$ are computed once before step 1 and kept fixed. At step $n$, the clip features $\vfeatb_n$ are concatenated with  $\sfeat$ along the feature dimension to get $\svfeat_n \in \mathbb{R}^{L \times (D_v + D_s)}$. The concatenated visual features $\svfeat_n$ and $\qfeat$ are used by the cross-modal encoder to perform joint reasoning: 
\begin{equation}
    \label{eqn:crossmodal_cs}
    \cfeatb_n = \textrm{CrossModalEncoder}(\svfeat_n, \qfeat).
\end{equation}
The selection policy is a two-layered MLP that predicts a binary value (i.e., thresholded probabilities) per clip indicating whether clip features ought to be computed or not:\footnote{Previously selected clips are excluded from predictions.}
\begin{equation}
    \label{eqn:binarypred}
    \bfeatb_{n+1} = \textrm{SelectionPolicy}(\cfeatb_n) \in \{0, 1\}^L.
\end{equation}
Finally, the video backbone is used to extract clip features for the clips selected in $\bfeatb_{n+1}$. They are added to $\vfeatb_n$ to get the updated clip features $\vfeatb_{n+1}$. This selection process is repeated for $N$ steps, and the cumulative set of clip features ($\vfeatb = \vfeatb_{N+1}$) is used to predict the EM response as shown in~\cref{eqn:crossmodal,eqn:localize}.

\subsection{RIO features for semantic indexing}
\label{sec:rio_feats}
Our SpotEM relies on cheap semantic indexing to preview the video and select clips intelligently. Prior work on efficient video recognition uses ImageNet-pretrained features for this purpose~\cite{wu2019adaframe,wu2019liteeval,gowda2021smart}. However, ImageNet features capture only object-level information, which is insufficient for query-conditioned indexing in EM, where contextual cues about human-object interactions and room-level characteristics are needed. For example, object interaction features may be useful for the query, \emph{``What tool did I use for fixing this part of the bike last time?"} while room-level features may be useful for the query, \emph{``Did I leave the lights on in the living room?"} 
Hence, we design a set of low-cost semantic indexing features that capture context from the \textbf{r}ooms/scenes visited, human-object \textbf{i}nteractions, and the visible \textbf{o}bjects, which we term \emph{RIO}. For each feature, we train an EfficientNet-b0 image encoder, which incurs 2-3 orders lower computational cost than typical clip encoders~\cite{tan2019efficientnet}. 

\textbf{Room features:} We train an image encoder as a room/scene classifier to capture scene characteristics using scene annotations for Ego4D videos~\cite{nagarajan2022egocentric}. This includes  28 categories of both indoor scenes (e.g., bedroom, living room) and outdoor scenes (e.g., garden, porch). 

\textbf{Interaction features:} We capture human-object interactions by training another image encoder using the contrastive vision-text pretraining objective from EgoVLP~\cite{lin2022egocentric}. The objective is to maximize feature similarity between \emph{video frames} and Ego4D's time-synchronized textual narrations reporting every step of the camera-wearer's activity~\cite{grauman2022ego4d}. In addition to image-text contrastive losses, we add losses to distill the clip-level features from a pretrained EgoVLP TimeSformer~\cite{bertasius2021space} backbone into the image encoder.

\textbf{Object features:} We replace the ImageNet features from prior work with self-supervised VICReg~\cite{bardes2022vicreg} features trained on Ego4D images. The objective in VICReg is to learn image representations in a self-supervised way by minimizing reconstruction errors between two different views of an image (invariance), while maintaining diversity over each feature dimension (variance), and decorrelating pairs of feature dimensions (covariance). This captures object properties and also bridges the visual domain gap experienced by ImageNet features. 

Overall, we sample one image within each video clip, extract each of the RIO features, and concatenate them to obtain the semantic indexing features $\sfeat$ described in~\cref{fig:nlq_arch}.

\subsection{Model optimization}
\label{sec:optim} 
Our SpotEM approach uses a MemorySpotter module for clip selection in conjunction with the EM modules, i.e., the cross-modal encoder and the localization module (see ~\cref{fig:nlq_arch}).  We jointly optimize these modules end-to-end on the EM task for improving task performance while only sampling a subset of video clips. We keep the video, semantic index, and text backbones frozen during training. Our loss function consists of the following terms: an EM task loss $\lossem$, a selection loss $\losssel$,  and two novel distillation losses for feature distillation $\lossfd$ and prediction distillation $\losspd$. 

\textbf{EM task loss} ($\lossem$) optimizes the model to improve the EM NLQ performance. It typically consists of prediction losses for start and end locations, and a loss to prioritize clips overlapping with the response~\cite{zhang2020span}. The details are in~\cref{suppsec:em_losses}.

\textbf{Selection loss} ($\losssel$) optimizes the model to select a specified budget of clips and penalizes under-/over-sampling: 
\begin{equation}
    \label{eqn:selection_loss}
    \losssel = \bigg(\mathbb{E}_{(\video,\query) \sim D_\textrm{train}} \bigg[\frac{1}{L}\sum_{l=1}^L \bfeatb_\textrm{joint}^l \bigg] - \gamma\bigg)^2,
\end{equation}
where $D_{train}$ is the training dataset and $\bar{\bm{b}}_\textrm{joint} = \sum_{n=1}^{N+1} \bfeatb_n$ is the overall binary selections after $N$ steps (refer to~\cref{eqn:binarypred}). $\losssel$ limits the fraction of clips selected, in expectation, to a predefined hyperparameter $\gamma$. This is similar to binary selection losses used in prior work~\cite{wu2019liteeval,meng2020ar}. We further regularize the per-step selection $\bfeatb_n^l$ by encouraging the model to select $(\frac{\gamma}{N})L$ clips in each step from $1$ to $N$. We ignore this term from~\cref{eqn:selection_loss} for brevity.  We observed this simple regularization can improve training stability. However, we did not explore more complex schemes that vary the number of selections conditioned on the iteration. Since MemorySpotter predicts binary selection values in~\cref{eqn:binarypred}, it is not differentiable for gradient-based optimization. Following prior work, we use the Gumbel-Softmax trick to reparameterize argmax sampling using a softmax relaxation during training~\cite{hazan2012partition,maddison2017concrete,jang2017categorical,wu2019liteeval}. See~\cref{suppsec:gumbel_max} for details.

\textbf{Distillation losses:} We observed some optimization difficulties during the joint optimization of MemorySpotter and EM modules (i.e., cross-modal encoder and localization module). Intuitively, the training of the EM modules is disrupted by the noisy clip-selection from MemorySpotter during the initial stage of learning. This in turn affects the optimization of MemorySpotter since it gets noisy gradients from the remaining modules. To break this negative feedback loop, we adopt a two-stage training pipeline. First, we train an expert EM model to perform the task without MemorySpotter (but include the semantic index). Then we use the expert to provide distillation supervision for joint optimization of student EM modules with MemorySpotter. 

For a given input $(\video, \query)$ and ground-truth response $\response$, let $\cfeatb_\textrm{expert}$ and $\cfeatb_\textrm{student}$ be the cross-modal encoder outputs from~\cref{eqn:crossmodal} for the expert and student EM models, respectively. Unlike the student, the expert uses all the video features. The \emph{feature distillation loss} $\lossfd$ trains the student to match the expert's cross-modal features. 
\begin{equation}
    \label{eqn:fd_loss}
    \lossfd = \big\|\textrm{StopGrad}(\cfeatb_\textrm{expert}) - \cfeatb_\textrm{student}\big\|_1,
\end{equation}
where the gradient is not propagated to the frozen expert. Similarly, we also define the \emph{prediction distillation loss} $\losspd$, which trains the student localization predictions to match the expert localization predictions. For example, in the VSLNet~\cite{zhang2020span}, this would train the student EM model to minimize the KL divergence between its predicted distribution over highlight scores, start and end locations, and the corresponding expert distributions. See~\cref{suppsec:em_losses} for more details. Overall, our final loss is:
\begin{equation}
    \label{eqn:loss_distill}
    \losstot = \lossem + \coefsel \losssel + \coeffd \lossfd + \coefpd \losspd,
\end{equation}
where the $\lambda_*$ are loss scaling hyperparameters determined via validation. Jointly, they encourage the model to improve EM performance while limiting the budget of clips selected. 

\begin{table}[t]
\resizebox{0.48\textwidth}{!}{
\begin{tabular}{@{}ccccccc@{}}
\toprule
 Row    &  Clip selection method              & $\eta$     & Sem. index     &     MR@1    &     MR@5    & TFLOPs    \\ \midrule
        &   ZeroClips                         & 100        &    ImageNet    &     3.85    &     8.64    &   0.1    \\ \midrule
   1    &   Random                            & 90         &    ImageNet    &     4.35    &     9.85    &  26.8    \\
   2    &   Uniform                           & 90         &    ImageNet    &     4.32    &     9.89    &  26.8    \\
   3    &   LiteEval~\cite{wu2019liteeval}    &90          &    ImageNet    &     5.82    &    11.53    &  26.6    \\
   4    &   OCSampler~\cite{lin2022ocsampler} &90          &    ImageNet    &     5.72    &    12.32    &  26.8    \\
   5    &   \modelname{} w/o distill          & 91         &    ImageNet    &     6.99    &    12.95    &  24.8    \\ 
   6    &   \modelname{} w/o distill          & 90         &      RIO       &     7.48    &    14.82    &  26.9    \\ 
   7    &   \modelname{} (ours)               & 90         &      RIO       & \tb{9.62}   &\tb{17.07}   &  27.0    \\ \midrule
   1    &   Random                            & 75         &    ImageNet    &     4.91    &    11.03    &  67.0    \\
   2    &   Uniform                           & 75         &    ImageNet    &     5.62    &    12.08    &  67.0    \\
   3    &   LiteEval~\cite{wu2019liteeval}    & 75         &    ImageNet    &     7.12    &    13.27    &  66.0    \\
   4    &   OCSampler~\cite{lin2022ocsampler} &75          &    ImageNet    &     7.34    &    14.49    &  67.0    \\
   5    &   \modelname{} w/o distill          & 76         &    ImageNet    &     9.22    &    16.90    &  65.2    \\
   6    &   \modelname{} w/o distill          & 75         &      RIO       &     9.92    &    17.27    &  66.9    \\ 
   7    &   \modelname{} (ours)               & 76         &      RIO       &\tb{11.06}   &\tb{18.92}   &  64.9    \\ \midrule
   1    &   Random                            & 50         &    ImageNet    &     7.52    &    14.85    & 133.9    \\
   2    &   Uniform                           & 50         &    ImageNet    &     8.24    &    15.75    & 133.9    \\
   3    &   LiteEval~\cite{wu2019liteeval}    & 50         &    ImageNet    &     8.70    &    16.21    & 132.8    \\
   4    &   OCSampler~\cite{lin2022ocsampler} &50          &    ImageNet    &     9.11    &    16.93    & 133.9    \\
   5    &   \modelname{} w/o distill          & 53         &    ImageNet    &     9.35    &    17.18    & 125.5    \\
   6    &   \modelname{} w/o distill          & 52         &      RIO       &     9.84    &    18.70    & 129.5    \\ 
   7    &   \modelname{} (ours)               & 51         &      RIO       &\tb{11.56}   &\tb{19.90}   & 131.9    \\ \midrule
 \rowcolor[HTML]{C0C0C0}
        &   AllClips~\cite{chen2022internvideo}& 0         &  -             &    11.45    &    20.56    & 267.6     \\ \bottomrule
\end{tabular}
}
\vspace*{-0.05in}
\caption{Comparing clip selection methods for the InternVideo EM method~\cite{chen2022internvideo} on the Ego4D NLQ benchmark.}
\label{tab:internvideo_results}
\end{table}

\begin{figure*}[t]
    \centering
    \includegraphics[width=1.0\textwidth,clip,trim={0 11cm 3cm 0}]{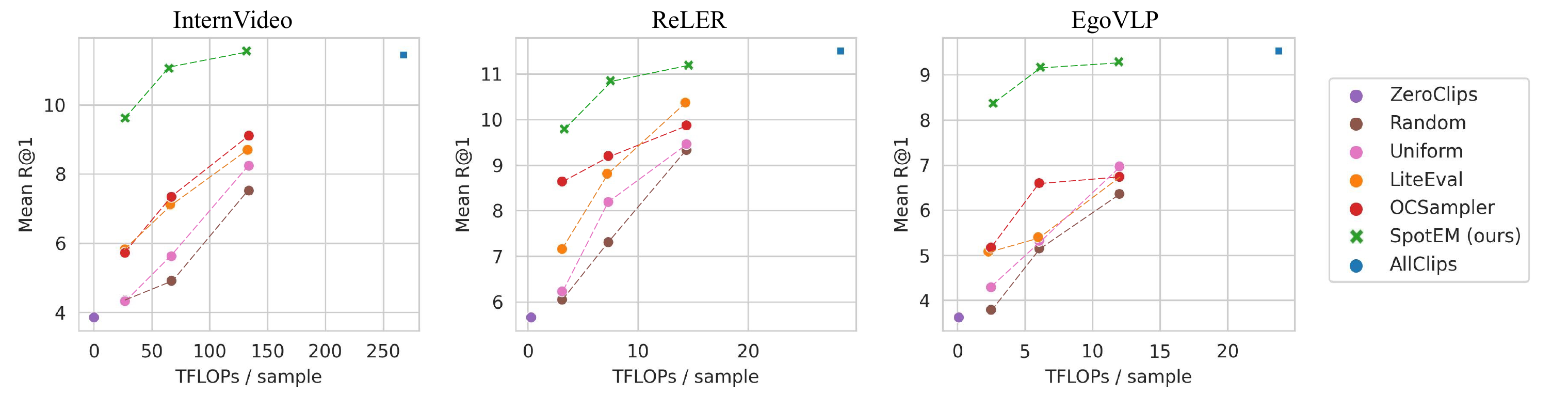}
    \vspace*{-0.25in}
    \caption{Plot comparing accuracy (y axis) vs. computational cost (x axis) for clip selection methods with InternVideo~\cite{chen2022internvideo}, ReLER~\cite{liu2022reler}, and EgoVLP~\cite{lin2022egocentric} EM methods. All methods are evaluated on the Ego4D NLQ benchmark. \modelname{} achieves over 84\% -- 97\% of the most expensive AllClips method with 4-10$\times$ lower computational cost. Note that the computation cost for each clip sampling method includes the cost of extracting semantic index features (see~\cref{suppsec:flops_calc} for more details). For each method, we show the complete results in~\cref{tab:internvideo_results,tab:reler_results,tab:egovlp_results}. 
    }
    \label{fig:plot_results}
    \vspace*{-0.1in}
\end{figure*}

\section{Experiments}

We next validate SpotEM's impact in practice.

\vspace*{-0.1in}
\subsection{Experimental setup}
\label{sec:exp_setup}
We evaluate our approach on the large-scale EM NLQ benchmark from Ego4D~\cite{grauman2022ego4d}, which is the only public dataset supporting this task to our knowledge. The dataset contains $11.3k / 3.9k / 4.0k$ queries annotated over $136 / 45 / 46$ hours of train/val/test videos. Each video clip is 8.2 minutes on average (with the longest video spanning 20 minutes), and each response window is 10.5 seconds on average. Unlike Ego4D, prior query localization and QA datasets focus only on third-person settings~\cite{regneri2013grounding,gao2017tall,tapaswi2016movieqa}, or derive videos from simulation with strong assumptions of ground-truth odometry~\cite{datta2022episodic}. 

While Ego4D uniquely supports the NLQ task by having long-form egocentric videos and natural language queries, we also test the generality of our approach by benchmarking it on the \tacos{} dataset for natural language grounding (NLG). Instead of localizing responses to natural language queries in egocentric videos like Ego4D NLQ, the goal in \tacos{} NLG is to localize short descriptions of the human activities in exocentric videos. It contains long exo videos of kitchen activities (5 minutes on average) and short natural language moments (5 seconds on average).

\textbf{Evaluation metrics:} We measure NLQ and NLG accuracy using MR@1 and MR@5 metrics, which are recall@\{1, 5\} averaged over temporal IoU values $[0.3, 0.5]$. We measure the savings in clip-feature computation using the efficiency-level metric:
\begin{equation}
    \eta = \mathbb{E}_{(V,Q) \sim D_{\textrm{val}}}[100 - k],
\end{equation}
where $k = 100\times\frac{\textrm{\scriptsize \# sampled clips}}{\textrm{\scriptsize \# total clips}}$. We quantify computational cost with TFLOPs measured using the DeepSpeed library~\cite{rasley2020deepspeed}. Since the clip feature extraction accounts for over 99.9\% of the computational cost,
TFLOPs is approximately inversely proportional to $\eta$.

\subsection{Baselines}
We perform experiments with multiple base EM methods and compare against several clip-selection baselines. In particular, we choose three state-of-the-art EM methods to demonstrate the effectiveness and versatility of our clip-selection approach.

\textbf{InternVideo}~\cite{chen2022internvideo} proposes a video foundation model that learns a single video representation to achieve state-of-the-art on several tasks including EM. This method won the ECCV 2022 EM challenge. 

\textbf{ReLER}~\cite{liu2022reler} uses a modified version of VSLNet-L~\cite{zhang2021natural} and proposes video-level data augmentation techniques for NLQ. This method was the winning entry in the CVPR 2022 EM challenge. 

\textbf{EgoVLP}~\cite{lin2022egocentric} performs large-scale egocentric video-language pretraining on paired (video clip, narrations text) from Ego4D. This method placed second in the CVPR 2022 EM challenge. 

For each base EM method, we compare \modelname{} against several clip sampling baselines that represent alternate ways to select clips for reducing the EM inference cost.

\textbf{ZeroClips:} This does not sample any clip features, and only uses the (cheap) semantic indexing features for EM. This serves as a lower bound for clip sampling methods and has the lowest computational cost. 

\textbf{Random:} This randomly samples $k\%$ of clips. 

\textbf{Uniform:} This samples $k\%$ of clips, uniformly spaced. 

\textbf{LiteEval:} This is our implementation of the model of~\citet{wu2019liteeval}, which linearly scans the video clips while making binary choices for clip selection. We modified it to include query features as additional inputs to the model. 

\textbf{OCSampler:} This is our implementation of the model of~\citet{lin2022ocsampler}, which previews each video clip independently, predicts per-clip selection scores and selects the top $k\%$ clips with the highest scores. We modified it to include query features as additional inputs. 

\textbf{AllClips:} This is the base EM model which performs the task using all video clips (i.e., no clip sampling). This serves as an upper bound for clip sampling methods and has the highest computational cost.

For all baselines other than AllClips, we use ImageNet semantic indexing features in addition to the selected clip features for performing EM. Please see~\cref{suppsec:implementation} for implementation details. 

\begin{figure*}[t]
    \centering
    \includegraphics[width=\textwidth,clip,trim={0 4.5cm 2.0cm 0}]{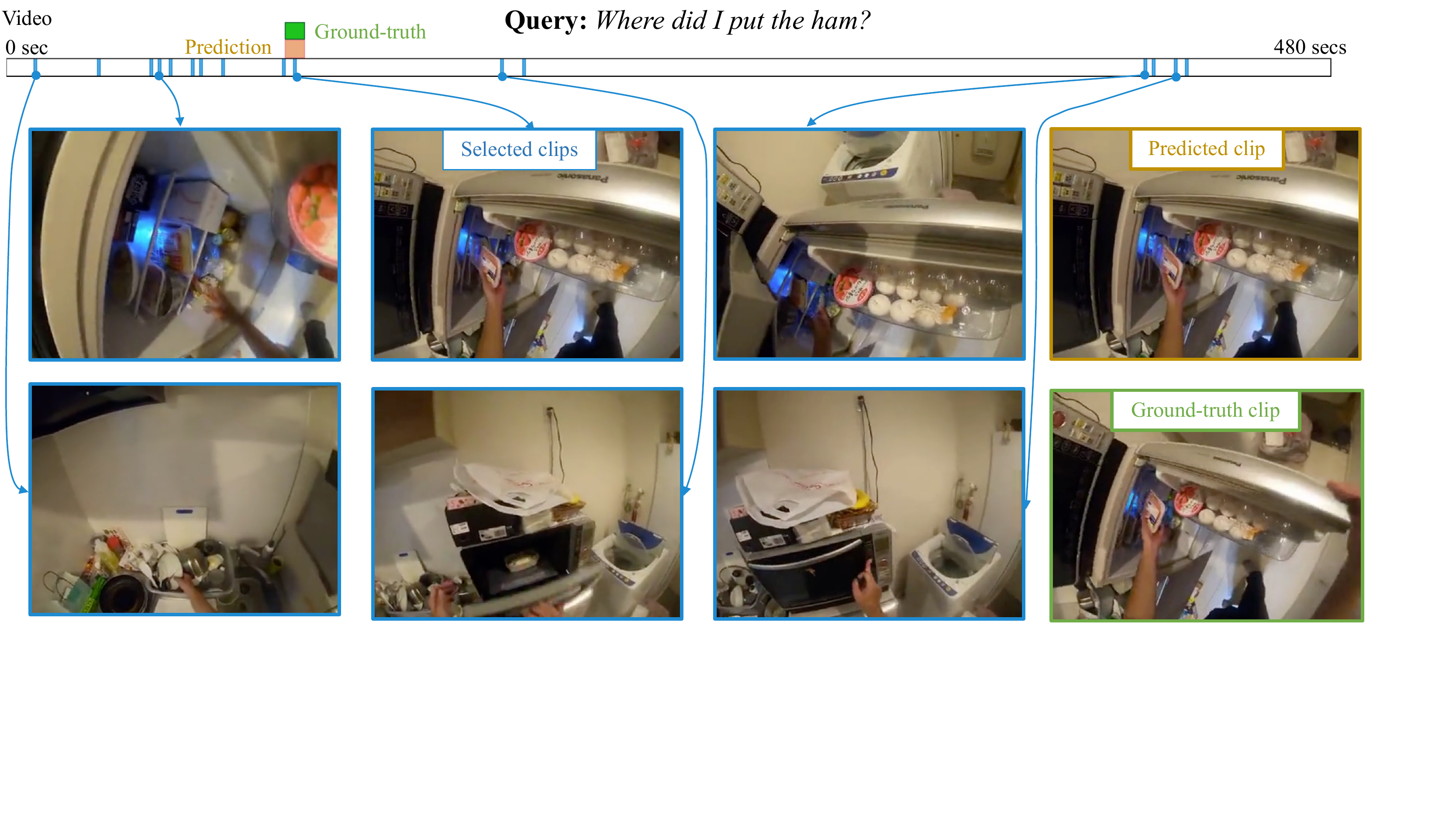}
    \vspace*{-0.2in}
    \caption{\small{We visualize an example of how \modelname{} efficiently performs EM. The query is shown on the top. Below the query, we show a temporal plot containing the video (8 minutes long) overlayed with the timestamps of the clips selected by the policy in light blue (a $\sim$0.5 seconds chunk of video).  The prediction (yellow) and ground truth (green) are shown above the video plot. For a representative set of clips selected by the model, we show the image taken from the center of the clip (highlighted in blue), followed by images from the center of the predicted (yellow) and ground-truth (green) time windows. \textbf{\modelname{}} uses RIO semantic index to preview the video cheaply, identifies clips that are likely to contain the interaction (put, ham), and correctly localizes the response. In~\cref{suppsec:qualitative}, we present more examples that highlight the effective clip selection strategies learned by SpotEM. We also present failure cases where SpotEM confuses closely related objects (e.g., salad dressing vs. salt container) and fine-grained attributes of objects (e.g., brown box vs. blue box).}
    }
    \label{fig:qualitative_2}
\end{figure*}

\subsection{Experimental results}
\label{sec:experiments}

In~\cref{tab:internvideo_results}, we compare \modelname{} with na\"ive baselines and prior state-of-the-art clip selection methods~\cite{wu2019liteeval,lin2022ocsampler} on the Ego4D NLQ benchmark. The base EM method is the state-of-the-art InternVideo method~\cite{chen2022internvideo}. We evaluate methods at different efficiency levels $\eta = [50, 75, 90]$. For random, uniform, and all learned methods, we train one model per efficiency level $\eta$. For random, uniform and OCSampler, this means sampling $k\% = 100 - \eta$ of clips during training and evaluation. For the LiteEval and \modelname{}, we set $\gamma = 1 - \frac{\eta}{100}$ in~\cref{eqn:selection_loss} during training.  Note that the efficiency for LiteEval and \modelname{} may be slightly larger than the specified $\eta$ since the selection is learned end-to-end without any manual intervention.

Consider rows 1-5 in~\cref{tab:internvideo_results}, which presents a comparison across all methods with ImageNet features as the semantic index. Among the baselines, we observe a consistent trend: 
\[
    \small \textrm{OCSampler} \ge \textrm{LiteEval} > \textrm{Uniform} \ge \textrm{Random} > \textrm{ZeroClips} 
\]
As expected, all methods are better than the lower-bound zero baseline. Uniform sampling captures information more effectively than random sampling. Importantly, all learned methods outperform na\"ive baselines by a good margin, confirming the value of intelligent clip selection for EM. 

Our \modelname{} in row 5 outperforms all the baselines, even before incorporating our RIO semantic index and distillation losses, showing the strength of our MemorySpotter architecture. When compared to the linear clip-wise scan strategy in LiteEval, our method previews the entire video using transformer-based attention and makes selection decisions for all clips at once. This results in 1 -- 3 higher absolute mean-recall (MR) than LiteEval. Our method also obtains 0.5 -- 2.5 higher absolute MR than OCSampler.  Unlike OCSampler, which performs one-shot score prediction for all clips, \modelname{} performs recursive clip selection by previewing the video multiple times. At each step, it selects a subset of video clips, extracts their (heavier) clip features, and uses this knowledge for future steps. This results in better performance, especially for higher efficiency levels.  This comparison is valuable to show how MemorySpotter's design overcomes limitations of SoTA sampling models designed for action recognition rather than episodic memory.

While the \modelname{} clip selection strategy from row 5 outperforms prior methods, it falls short of the original InternVideo results (NB: the latter uses dramatically more computation). However, when we replace the ImageNet features with our RIO features in row 6, we observe good improvements of 0.5 -- 2 MR with little to no increase in the computational cost. This confirms the value of building a (cheap) semantic index that captures room, interaction, and object features for efficient EM. When we further add distillation losses to improve model optimization in row 7, the performance improves by 1.5 -- 2 MR and effectively bridges the gap between our efficient \modelname{} and the base (status quo) 
EM method which uses all clips. Specifically, we can achieve 84\%, 96.5\% and 100\% of original MR@1 metric with 10$\times$, 4$\times$ and 2$\times$ reduction in computational cost, respectively.  By combining our novel clip selection model, RIO semantic index, and distillation losses during training, \modelname{} successfully reduces the computational cost by 4$\times$ while maintaining over $95\%$ of the EM performance. Please see~\cref{fig:plot_results} (left) for a performance vs. cost chart summarizing our findings. We present a qualitative analysis of \modelname{} in~\cref{fig:qualitative_2}.

We also test \modelname{} on two more EM models: ReLER~\cite{liu2022reler} in \cref{fig:plot_results} (center) and EgoVLP~\cite{lin2022egocentric} in \cref{fig:plot_results} (right). In both cases, \modelname{} outperforms all baselines across efficiency levels, and attains at least 95\% of the original EM method's accuracy with a 4$\times$ reduction in computational cost. See~\cref{suppsec:full_results} for results analogous to~\cref{tab:internvideo_results} for ReLER and EgoVLP. The trends are similar as those for InternVideo above, confirming that our approach generalizes across multiple EM methods. 

In~\cref{suppsec:tacos_results}, we compare clip sampling methods on the \tacos{} dataset for natural language grounding. The trends largely echo our results on the Ego4D NLQ benchmark and confirm the advantages of SpotEM for NLG on exo videos. However, we find that the RIO features are not beneficial, likely due to the ego-exo domain shift since RIO features are trained on egocentric images.

\subsection{Ablation studies}
Thus far, we compared \modelname{} with several clip-selection baselines and demonstrated its superiority. We now present ablation studies on the Ego4D NLQ benchmark to analyze different aspects of our model.

\textbf{Ablation of RIO semantic index:} In~\cref{sec:experiments}, we confirmed the benefits of using RIO features over standard ImageNet features. We now study the impact of removing each RIO feature, one by one, on the EM task (see~\cref{tab:ablation_feats}). The base EM method is InternVideo. We study \modelname{} without distillation losses to avoid exhaustively training expert models for each feature set. At $\eta=50\%$, the impact of removing any one feature is minimal since 50\% of the expensive clip features are already available. At higher efficiencies, the performance noticeably decreases when we remove any one feature. This study confirms our intuitions: RIO semantic index captures complementary aspects of the EM task and are critical to spotting query-relevant clips. 

In~\cref{suppsec:rio_ablation}, we analyze the impact of removing RIO features across query templates (e.g., \emph{where is object X? what did I put in X? what is X before/after event Y?}) and study the role of RIO features for performing EM efficiently. Our study indicates that each of the RIO features has a varying impact on the templates. Removing room features affects queries with strong scene association (e.g., ovens are in kitchens), while removing interaction features affects queries that require object-interaction reasoning (e.g., \emph{``how many drawers did I open?"}). We also found that RIO features facilitate intelligent clip-selection, but do not improve the absolute EM performance themselves.

\begin{figure}[t]
    \centering
    \includegraphics[width=0.48\textwidth,clip,trim={0 12cm 11cm 0}]{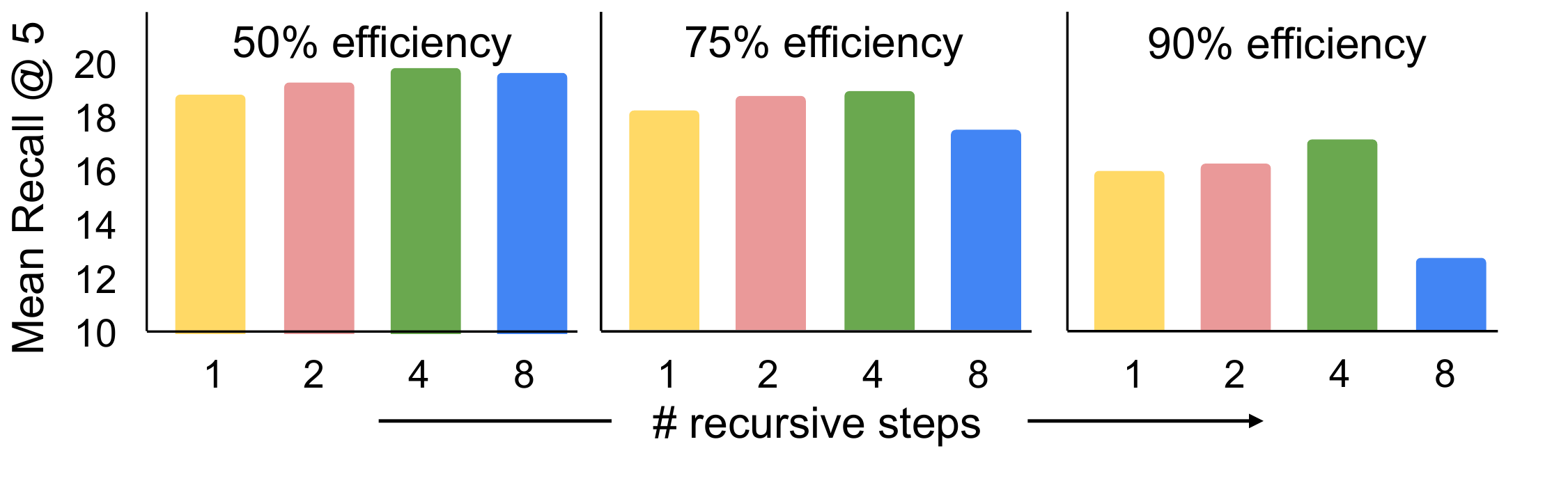}
    \vspace*{-0.3in}
    \caption{\textbf{Ablation study of \# recursive steps:} We assess the impact of the recursive step size $N$ for \modelname{} on MR@1. }
    \label{fig:ablation_steps}
\end{figure}

\textbf{Ablation of recursive clip selection:} Unlike prior methods by~\citet{wu2019liteeval} and~\citet{lin2022ocsampler} that were designed to handle shorter videos for recognition, we explicitly target longer videos for EM using our recursive clip-selection approach, where we preview the entire video, and actively select subsets of clips over multiple steps. We assess the impact of the recursion length $N$ in~\cref{fig:ablation_steps}. Specifically, we train \modelname{} models with $\eta=[50,75,90]$ and $N=[1,2,4,8]$, where $N=4$ is our default choice. $N=1$ is the non-recursive case where the model looks at only the semantic index for selecting all clips, without incorporating any heavy clip features for selection. Across efficiency levels, we observe that the mean recall@5 increases from $N=1$ to $N=4$, confirming the value of recursively selecting clips and incorporating knowledge from prior clips for future selection. The performance reduces beyond $N=4$ since gradient propagation becomes more challenging.

We further analyze SpotEM's performance as a function of video duration in~\cref{suppsec:video_duration} and find that it continues to perform well on long videos, outperforming the baseline sampling methods. \modelname{} also achieves $85\%$ of the base EM method's performance, while sampling only $25\%$ of the clips. However, the absolute performance is ultimately limited by the underlying EM method. We also present a detailed study of SpotEM's clip selection behaviors in~\cref{suppsec:spotem_behavior}. Our experiments verify our intuition that SpotEM samples query-relevant clips that are useful for the base EM method to respond to the query. \modelname{} does not try to respond to the query directly on its own.

\begin{table}[t]
\centering
\resizebox{0.42\textwidth}{!}{
\begin{tabular}{@{}ccccccccc@{}}
\toprule

     &     &      &   \multicolumn{3}{c}{MR@1 at $\eta$}       &   \multicolumn{3}{c}{MR@5 at $\eta$}         \\ \cmidrule(lr){4-6} \cmidrule(lr){7-9}
 R  & I    &  O    &     50        &     75      &     90       &     50        &    75       &     90         \\ \midrule
\yes& \yes &  \yes &     9.84      &\tb{9.92}    &\tb{7.48}     &\tb{18.70}     &\tb{17.27}   & \tb{16.83}     \\
    & \yes &  \yes &     9.83      &    9.26     &    7.26      &    17.38      &    17.08    &     13.36      \\
\yes&      &  \yes &     9.52      &    8.36     &    5.89      &    17.23      &    14.83    &     11.05      \\
\yes& \yes &       & \tb{10.38}    &    9.04     &    6.94      &    18.53      &    16.30    &     13.12      \\
\bottomrule
\end{tabular}
}
\vspace*{-0.05in}
\caption{\small \textbf{Ablation study of semantic index:} We assess the impact of removing each one of the RIO features on EM performance of InternVideo~\cite{chen2022internvideo}. The first 3 columns indicate whether object, room, interaction features are used, respectively. The second row shows the efficiency level $\eta$ of the model.}
\label{tab:ablation_feats}
\vspace*{-0.1in}
\end{table}

\section{Conclusions}

\modelname{} tackles the efficiency challenge of answering episodic memory queries on long egocentric video head-on.  Our novel MemorySpotter clip selection policy, learned jointly with the EM  model, leverages cheap video features to prioritize the temporal regions most likely to yield a given natural language query's answer.  Together with our novel distillation losses and well-designed RIO semantic indexing features, \modelname{} successfully reduces the cost for multiple SoTA EM methods---4 to 10$\times$ efficiency gains while maintaining high accuracy.  While the videos in our study already represent a sizeable jump in length compared to mainstream video understanding work, in future work, we are interested in exploring a hierarchical variant of our model to triage the video across hours of content.

\section{Acknowledgements}
UT Austin is supported in part by the IFML NSF AI Institute and NSF CCRI. KG is a paid as a researcher at Meta. We thank the ICML reviewers and area chair for their valuable feedback.

\bibliography{main}

\begin{thebibliography}{67}
\providecommand{\natexlab}[1]{#1}
\providecommand{\url}[1]{\texttt{#1}}
\expandafter\ifx\csname urlstyle\endcsname\relax
  \providecommand{\doi}[1]{doi: #1}\else
  \providecommand{\doi}{doi: \begingroup \urlstyle{rm}\Url}\fi

\bibitem[Abu~Farha et~al.(2018)Abu~Farha, Richard, and Gall]{abu2018will}
Abu~Farha, Y., Richard, A., and Gall, J.
\newblock When will you do what?-anticipating temporal occurrences of
  activities.
\newblock In \emph{Proceedings of the IEEE conference on computer vision and
  pattern recognition}, pp.\  5343--5352, 2018.

\bibitem[Bardes et~al.(2022)Bardes, Ponce, and LeCun]{bardes2022vicreg}
Bardes, A., Ponce, J., and LeCun, Y.
\newblock Vicreg: Variance-invariance-covariance regularization for
  self-supervised learning.
\newblock In \emph{ICLR}, 2022.

\bibitem[Bertasius et~al.(2021)Bertasius, Wang, and
  Torresani]{bertasius2021space}
Bertasius, G., Wang, H., and Torresani, L.
\newblock Is space-time attention all you need for video understanding?
\newblock In \emph{ICML}, volume~2, pp.\ ~4, 2021.

\bibitem[Caba~Heilbron et~al.(2015)Caba~Heilbron, Escorcia, Ghanem, and
  Carlos~Niebles]{caba2015activitynet}
Caba~Heilbron, F., Escorcia, V., Ghanem, B., and Carlos~Niebles, J.
\newblock Activitynet: A large-scale video benchmark for human activity
  understanding.
\newblock In \emph{Proceedings of the ieee conference on computer vision and
  pattern recognition}, pp.\  961--970, 2015.

\bibitem[Cai et~al.(2018)Cai, Kitani, and Sato]{cai2018understanding}
Cai, M., Kitani, K., and Sato, Y.
\newblock Understanding hand-object manipulation by modeling the contextual
  relationship between actions, grasp types and object attributes.
\newblock \emph{arXiv preprint arXiv:1807.08254}, 2018.

\bibitem[Chen et~al.(2011)Chen, Bilgic, Getoor, and Jacobs]{chen2011dynamic}
Chen, D., Bilgic, M., Getoor, L., and Jacobs, D.
\newblock Dynamic processing allocation in video.
\newblock \emph{IEEE transactions on pattern analysis and machine
  intelligence}, 33\penalty0 (11):\penalty0 2174--2187, 2011.

\bibitem[Chen et~al.(2022)Chen, Xing, Chen, Wang, Li, Li, Liu, Wang, Zheng,
  Huang, et~al.]{chen2022internvideo}
Chen, G., Xing, S., Chen, Z., Wang, Y., Li, K., Li, Y., Liu, Y., Wang, J.,
  Zheng, Y.-D., Huang, B., et~al.
\newblock Internvideo-ego4d: A pack of champion solutions to ego4d challenges.
\newblock \emph{arXiv preprint arXiv:2211.09529}, 2022.

\bibitem[Chen et~al.(2018)Chen, Wang, Zhang, and Huang]{chen2018less}
Chen, Y., Wang, S., Zhang, W., and Huang, Q.
\newblock Less is more: Picking informative frames for video captioning.
\newblock In \emph{Proceedings of the European conference on computer vision
  (ECCV)}, pp.\  358--373, 2018.

\bibitem[Damen et~al.(2014)Damen, Leelasawassuk, Haines, Calway, and
  Mayol-Cuevas]{damen2014you}
Damen, D., Leelasawassuk, T., Haines, O., Calway, A., and Mayol-Cuevas, W.~W.
\newblock You-do, i-learn: Discovering task relevant objects and their modes of
  interaction from multi-user egocentric video.
\newblock In \emph{BMVC}, volume~2, pp.\ ~3, 2014.

\bibitem[Damen et~al.(2022)Damen, Doughty, Farinella, , Furnari, Ma, Kazakos,
  Moltisanti, Munro, Perrett, Price, and Wray]{Damen2022RESCALING}
Damen, D., Doughty, H., Farinella, G.~M., , Furnari, A., Ma, J., Kazakos, E.,
  Moltisanti, D., Munro, J., Perrett, T., Price, W., and Wray, M.
\newblock Rescaling egocentric vision: Collection, pipeline and challenges for
  epic-kitchens-100.
\newblock \emph{International Journal of Computer Vision (IJCV)}, 130:\penalty0
  33–55, 2022.
\newblock URL \url{https://doi.org/10.1007/s11263-021-01531-2}.

\bibitem[Datta et~al.(2022)Datta, Dharur, Cartillier, Desai, Khanna, Batra, and
  Parikh]{datta2022episodic}
Datta, S., Dharur, S., Cartillier, V., Desai, R., Khanna, M., Batra, D., and
  Parikh, D.
\newblock Episodic memory question answering.
\newblock In \emph{Proceedings of the IEEE/CVF Conference on Computer Vision
  and Pattern Recognition}, pp.\  19119--19128, 2022.

\bibitem[Del~Molino et~al.(2016)Del~Molino, Tan, Lim, and
  Tan]{del2016summarization}
Del~Molino, A.~G., Tan, C., Lim, J.-H., and Tan, A.-H.
\newblock Summarization of egocentric videos: A comprehensive survey.
\newblock \emph{IEEE Transactions on Human-Machine Systems}, 47\penalty0
  (1):\penalty0 65--76, 2016.

\bibitem[Fan et~al.(2018)Fan, Xu, Zhu, Yan, Ge, and Yang]{fan2018watching}
Fan, H., Xu, Z., Zhu, L., Yan, C., Ge, J., and Yang, Y.
\newblock Watching a small portion could be as good as watching all: Towards
  efficient video classification.
\newblock In \emph{Proceedings of the Twenty-Seventh International Joint
  Conference on Artificial Intelligence, {IJCAI-18}}, pp.\  705--711.
  International Joint Conferences on Artificial Intelligence Organization, 7
  2018.
\newblock \doi{10.24963/ijcai.2018/98}.
\newblock URL \url{https://doi.org/10.24963/ijcai.2018/98}.

\bibitem[Fathi et~al.(2011)Fathi, Ren, and Rehg]{fathi2011gtea}
Fathi, A., Ren, X., and Rehg, J.~M.
\newblock Learning to recognize objects in egocentric activities.
\newblock In \emph{CVPR 2011}, pp.\  3281--3288, 2011.
\newblock \doi{10.1109/CVPR.2011.5995444}.

\bibitem[Feichtenhofer(2020)]{feichtenhofer2020x3d}
Feichtenhofer, C.
\newblock X3d: Expanding architectures for efficient video recognition.
\newblock In \emph{Proceedings of the IEEE/CVF Conference on Computer Vision
  and Pattern Recognition}, pp.\  203--213, 2020.

\bibitem[Feichtenhofer et~al.(2019)Feichtenhofer, Fan, Malik, and
  He]{feichtenhofer2019slowfast}
Feichtenhofer, C., Fan, H., Malik, J., and He, K.
\newblock Slowfast networks for video recognition.
\newblock In \emph{Proceedings of the IEEE/CVF international conference on
  computer vision}, pp.\  6202--6211, 2019.

\bibitem[Furnari \& Farinella(2020)Furnari and Farinella]{furnari2020rolling}
Furnari, A. and Farinella, G.~M.
\newblock Rolling-unrolling lstms for action anticipation from first-person
  video.
\newblock \emph{IEEE transactions on pattern analysis and machine
  intelligence}, 43\penalty0 (11):\penalty0 4021--4036, 2020.

\bibitem[Furnari et~al.(2016)Furnari, Farinella, and
  Battiato]{furnari2016recognizing}
Furnari, A., Farinella, G.~M., and Battiato, S.
\newblock Recognizing personal locations from egocentric videos.
\newblock \emph{IEEE Transactions on Human-Machine Systems}, 47\penalty0
  (1):\penalty0 6--18, 2016.

\bibitem[Gao et~al.(2017)Gao, Sun, Yang, and Nevatia]{gao2017tall}
Gao, J., Sun, C., Yang, Z., and Nevatia, R.
\newblock Tall: Temporal activity localization via language query.
\newblock In \emph{Proceedings of the IEEE international conference on computer
  vision}, pp.\  5267--5275, 2017.

\bibitem[Gao et~al.(2020)Gao, Oh, Grauman, and Torresani]{gao2020listen}
Gao, R., Oh, T.-H., Grauman, K., and Torresani, L.
\newblock Listen to look: Action recognition by previewing audio.
\newblock In \emph{Proceedings of the IEEE/CVF Conference on Computer Vision
  and Pattern Recognition (CVPR)}, June 2020.

\bibitem[Girdhar \& Grauman(2021)Girdhar and Grauman]{girdhar2021anticipative}
Girdhar, R. and Grauman, K.
\newblock Anticipative video transformer.
\newblock In \emph{Proceedings of the IEEE/CVF International Conference on
  Computer Vision}, pp.\  13505--13515, 2021.

\bibitem[Girdhar et~al.(2022)Girdhar, Singh, Ravi, van~der Maaten, Joulin, and
  Misra]{girdhar2022omnivore}
Girdhar, R., Singh, M., Ravi, N., van~der Maaten, L., Joulin, A., and Misra, I.
\newblock Omnivore: A single model for many visual modalities.
\newblock In \emph{Proceedings of the IEEE/CVF Conference on Computer Vision
  and Pattern Recognition}, pp.\  16102--16112, 2022.

\bibitem[Gowda et~al.(2021)Gowda, Rohrbach, and Sevilla-Lara]{gowda2021smart}
Gowda, S.~N., Rohrbach, M., and Sevilla-Lara, L.
\newblock Smart frame selection for action recognition.
\newblock In \emph{Proceedings of the AAAI Conference on Artificial
  Intelligence}, volume~35, pp.\  1451--1459, 2021.

\bibitem[Grauman et~al.(2022)Grauman, Westbury, Byrne, Chavis, Furnari,
  Girdhar, Hamburger, Jiang, Liu, Liu, et~al.]{grauman2022ego4d}
Grauman, K., Westbury, A., Byrne, E., Chavis, Z., Furnari, A., Girdhar, R.,
  Hamburger, J., Jiang, H., Liu, M., Liu, X., et~al.
\newblock Ego4d: Around the world in 3,000 hours of egocentric video.
\newblock In \emph{Proceedings of the IEEE/CVF Conference on Computer Vision
  and Pattern Recognition}, pp.\  18995--19012, 2022.

\bibitem[Hazan \& Jaakkola(2012)Hazan and Jaakkola]{hazan2012partition}
Hazan, T. and Jaakkola, T.~S.
\newblock On the partition function and random maximum a-posteriori
  perturbations.
\newblock In \emph{ICML}, 2012.

\bibitem[Hou et~al.(2022)Hou, Zhong, Ji, Gao, Yan, Chan, Ngo, Shou, and
  Duan]{hou2022efficient}
Hou, Z., Zhong, W., Ji, L., Gao, D., Yan, K., Chan, W.-K., Ngo, C.-W., Shou,
  Z., and Duan, N.
\newblock An efficient coarse-to-fine alignment framework@ ego4d natural
  language queries challenge 2022.
\newblock \emph{arXiv preprint arXiv:2211.08776}, 2022.

\bibitem[Jang et~al.(2017)Jang, Gu, and Poole]{jang2017categorical}
Jang, E., Gu, S., and Poole, B.
\newblock Categorical reparametrization with gumble-softmax.
\newblock In \emph{International Conference on Learning Representations (ICLR
  2017)}. OpenReview. net, 2017.

\bibitem[Jiang et~al.(2015)Jiang, Dai, Mei, Rui, and Chang]{jiang2015super}
Jiang, Y.-G., Dai, Q., Mei, T., Rui, Y., and Chang, S.-F.
\newblock Super fast event recognition in internet videos.
\newblock \emph{IEEE Transactions on Multimedia}, 17\penalty0 (8):\penalty0
  1174--1186, 2015.
\newblock \doi{10.1109/TMM.2015.2436813}.

\bibitem[Jiang et~al.(2018)Jiang, Wu, Wang, Xue, and Chang]{FCVID}
Jiang, Y.-G., Wu, Z., Wang, J., Xue, X., and Chang, S.-F.
\newblock Exploiting feature and class relationships in video categorization
  with regularized deep neural networks.
\newblock \emph{{IEEE} Transactions on Pattern Analysis and Machine
  Intelligence}, 40\penalty0 (2):\penalty0 352--364, 2018.
\newblock \doi{10.1109/TPAMI.2017.2670560}.
\newblock URL \url{https://doi.org/10.1109/TPAMI.2017.2670560}.

\bibitem[Kazakos et~al.(2019)Kazakos, Nagrani, Zisserman, and
  Damen]{kazakos2019epic}
Kazakos, E., Nagrani, A., Zisserman, A., and Damen, D.
\newblock Epic-fusion: Audio-visual temporal binding for egocentric action
  recognition.
\newblock In \emph{Proceedings of the IEEE/CVF International Conference on
  Computer Vision}, pp.\  5492--5501, 2019.

\bibitem[Korbar et~al.(2019)Korbar, Tran, and Torresani]{korbar2019scsampler}
Korbar, B., Tran, D., and Torresani, L.
\newblock Scsampler: Sampling salient clips from video for efficient action
  recognition.
\newblock In \emph{Proceedings of the IEEE/CVF International Conference on
  Computer Vision}, pp.\  6232--6242, 2019.

\bibitem[Lee \& Grauman(2015)Lee and Grauman]{yongjae-ijcv2015}
Lee, Y.~J. and Grauman, K.
\newblock Predicting important objects for egocentric video summarization.
\newblock \emph{International Journal on Computer Vision}, 2015.

\bibitem[Li et~al.(2021)Li, Wu, Shrivastava, and Davis]{li20212d}
Li, H., Wu, Z., Shrivastava, A., and Davis, L.~S.
\newblock 2d or not 2d? adaptive 3d convolution selection for efficient video
  recognition.
\newblock In \emph{Proceedings of the IEEE/CVF Conference on Computer Vision
  and Pattern Recognition}, pp.\  6155--6164, 2021.

\bibitem[Lin et~al.(2022{\natexlab{a}})Lin, Duan, Chen, Lin, and
  Wang]{lin2022ocsampler}
Lin, J., Duan, H., Chen, K., Lin, D., and Wang, L.
\newblock Ocsampler: Compressing videos to one clip with single-step sampling.
\newblock In \emph{Proceedings of the IEEE/CVF Conference on Computer Vision
  and Pattern Recognition}, pp.\  13894--13903, 2022{\natexlab{a}}.

\bibitem[Lin et~al.(2022{\natexlab{b}})Lin, Wang, Soldan, Wray, Yan, Xu, Gao,
  Tu, Zhao, Kong, et~al.]{lin2022egocentric}
Lin, K.~Q., Wang, A.~J., Soldan, M., Wray, M., Yan, R., Xu, E.~Z., Gao, D., Tu,
  R., Zhao, W., Kong, W., et~al.
\newblock Egocentric video-language pretraining.
\newblock \emph{arXiv preprint arXiv:2206.01670}, 2022{\natexlab{b}}.

\bibitem[Liu et~al.(2022)Liu, Wang, Li, Yang, and Zhuang]{liu2022reler}
Liu, N., Wang, X., Li, X., Yang, Y., and Zhuang, Y.
\newblock Reler@ zju-alibaba submission to the ego4d natural language queries
  challenge 2022.
\newblock \emph{arXiv preprint arXiv:2207.00383}, 2022.

\bibitem[Maddison et~al.(2017)Maddison, Mnih, and Teh]{maddison2017concrete}
Maddison, C., Mnih, A., and Teh, Y.
\newblock The concrete distribution: A continuous relaxation of discrete random
  variables.
\newblock In \emph{Proceedings of the international conference on learning
  Representations}. International Conference on Learning Representations, 2017.

\bibitem[Meng et~al.(2020)Meng, Lin, Panda, Sattigeri, Karlinsky, Oliva,
  Saenko, and Feris]{meng2020ar}
Meng, Y., Lin, C.-C., Panda, R., Sattigeri, P., Karlinsky, L., Oliva, A.,
  Saenko, K., and Feris, R.
\newblock Ar-net: Adaptive frame resolution for efficient action recognition.
\newblock In \emph{European Conference on Computer Vision}, pp.\  86--104.
  Springer, 2020.

\bibitem[Mo et~al.(2022)Mo, Mu, and Li]{mo2022simple}
Mo, S., Mu, F., and Li, Y.
\newblock A simple transformer-based model for ego4d natural language queries
  challenge.
\newblock \emph{arXiv preprint arXiv:2211.08704}, 2022.

\bibitem[Nagarajan et~al.(2020)Nagarajan, Li, Feichtenhofer, and
  Grauman]{nagarajan2020ego}
Nagarajan, T., Li, Y., Feichtenhofer, C., and Grauman, K.
\newblock Ego-topo: Environment affordances from egocentric video.
\newblock In \emph{Proceedings of the IEEE/CVF Conference on Computer Vision
  and Pattern Recognition}, pp.\  163--172, 2020.

\bibitem[Nagarajan et~al.(2022)Nagarajan, Ramakrishnan, Desai, Hillis, and
  Grauman]{nagarajan2022egocentric}
Nagarajan, T., Ramakrishnan, S.~K., Desai, R., Hillis, J., and Grauman, K.
\newblock Egocentric scene context for human-centric environment understanding
  from video.
\newblock \emph{arXiv preprint arXiv:2207.11365}, 2022.

\bibitem[Ortis et~al.(2017)Ortis, Farinella, D’Amico, Addesso, Torrisi, and
  Battiato]{ortis2017organizing}
Ortis, A., Farinella, G.~M., D’Amico, V., Addesso, L., Torrisi, G., and
  Battiato, S.
\newblock Organizing egocentric videos of daily living activities.
\newblock \emph{Pattern Recognition}, 72:\penalty0 207--218, 2017.

\bibitem[Panda et~al.(2021)Panda, Chen, Fan, Sun, Saenko, Oliva, and
  Feris]{panda2021adamml}
Panda, R., Chen, C.-F.~R., Fan, Q., Sun, X., Saenko, K., Oliva, A., and Feris,
  R.
\newblock Adamml: Adaptive multi-modal learning for efficient video
  recognition.
\newblock In \emph{Proceedings of the IEEE/CVF International Conference on
  Computer Vision (ICCV)}, pp.\  7576--7585, October 2021.

\bibitem[Price et~al.(2022)Price, Vondrick, and Damen]{price2022unweavenet}
Price, W., Vondrick, C., and Damen, D.
\newblock Unweavenet: Unweaving activity stories.
\newblock In \emph{Proceedings of the IEEE/CVF Conference on Computer Vision
  and Pattern Recognition}, pp.\  13770--13779, 2022.

\bibitem[Radford et~al.(2021)Radford, Kim, Hallacy, Ramesh, Goh, Agarwal,
  Sastry, Askell, Mishkin, Clark, et~al.]{radford2021learning}
Radford, A., Kim, J.~W., Hallacy, C., Ramesh, A., Goh, G., Agarwal, S., Sastry,
  G., Askell, A., Mishkin, P., Clark, J., et~al.
\newblock Learning transferable visual models from natural language
  supervision.
\newblock In \emph{International Conference on Machine Learning}, pp.\
  8748--8763. PMLR, 2021.

\bibitem[Rasley et~al.(2020)Rasley, Rajbhandari, Ruwase, and
  He]{rasley2020deepspeed}
Rasley, J., Rajbhandari, S., Ruwase, O., and He, Y.
\newblock Deepspeed: System optimizations enable training deep learning models
  with over 100 billion parameters.
\newblock In \emph{Proceedings of the 26th ACM SIGKDD International Conference
  on Knowledge Discovery \& Data Mining}, KDD '20, pp.\  3505–3506, New York,
  NY, USA, 2020. Association for Computing Machinery.
\newblock ISBN 9781450379984.
\newblock \doi{10.1145/3394486.3406703}.
\newblock URL \url{https://doi.org/10.1145/3394486.3406703}.

\bibitem[Regneri et~al.(2013)Regneri, Rohrbach, Wetzel, Thater, Schiele, and
  Pinkal]{regneri2013grounding}
Regneri, M., Rohrbach, M., Wetzel, D., Thater, S., Schiele, B., and Pinkal, M.
\newblock Grounding action descriptions in videos.
\newblock \emph{Transactions of the Association for Computational Linguistics},
  1:\penalty0 25--36, 2013.

\bibitem[Rohrbach et~al.(2014)Rohrbach, Rohrbach, Qiu, Friedrich, Pinkal, and
  Schiele]{rohrbach2014coherent}
Rohrbach, A., Rohrbach, M., Qiu, W., Friedrich, A., Pinkal, M., and Schiele, B.
\newblock Coherent multi-sentence video description with variable level of
  detail.
\newblock In \emph{Pattern Recognition: 36th German Conference, GCPR 2014,
  M{\"u}nster, Germany, September 2-5, 2014, Proceedings 36}, pp.\  184--195.
  Springer, 2014.

\bibitem[Rohrbach et~al.(2017)Rohrbach, Torabi, Rohrbach, Tandon, Pal,
  Larochelle, Courville, and Schiele]{rohrbach2017movie}
Rohrbach, A., Torabi, A., Rohrbach, M., Tandon, N., Pal, C., Larochelle, H.,
  Courville, A., and Schiele, B.
\newblock Movie description.
\newblock \emph{International Journal of Computer Vision}, 123\penalty0
  (1):\penalty0 94--120, 2017.

\bibitem[Sanh et~al.(2019)Sanh, Debut, Chaumond, and Wolf]{sanh2019distilbert}
Sanh, V., Debut, L., Chaumond, J., and Wolf, T.
\newblock Distilbert, a distilled version of bert: smaller, faster, cheaper and
  lighter.
\newblock \emph{arXiv preprint arXiv:1910.01108}, 2019.

\bibitem[Seo et~al.(2016)Seo, Kembhavi, Farhadi, and
  Hajishirzi]{seo2016bidirectional}
Seo, M., Kembhavi, A., Farhadi, A., and Hajishirzi, H.
\newblock Bidirectional attention flow for machine comprehension.
\newblock \emph{arXiv preprint arXiv:1611.01603}, 2016.

\bibitem[Suin \& Rajagopalan(2020)Suin and Rajagopalan]{suin2020efficient}
Suin, M. and Rajagopalan, A.
\newblock An efficient framework for dense video captioning.
\newblock In \emph{Proceedings of the AAAI Conference on Artificial
  Intelligence}, volume~34, pp.\  12039--12046, 2020.

\bibitem[Tan \& Le(2019)Tan and Le]{tan2019efficientnet}
Tan, M. and Le, Q.
\newblock Efficientnet: Rethinking model scaling for convolutional neural
  networks.
\newblock In \emph{International conference on machine learning}, pp.\
  6105--6114. PMLR, 2019.

\bibitem[Tapaswi et~al.(2016)Tapaswi, Zhu, Stiefelhagen, Torralba, Urtasun, and
  Fidler]{tapaswi2016movieqa}
Tapaswi, M., Zhu, Y., Stiefelhagen, R., Torralba, A., Urtasun, R., and Fidler,
  S.
\newblock Movieqa: Understanding stories in movies through question-answering.
\newblock In \emph{Proceedings of the IEEE conference on computer vision and
  pattern recognition}, pp.\  4631--4640, 2016.

\bibitem[Tong et~al.(2022)Tong, Song, Wang, and Wang]{tong2022videomae}
Tong, Z., Song, Y., Wang, J., and Wang, L.
\newblock Videomae: Masked autoencoders are data-efficient learners for
  self-supervised video pre-training.
\newblock \emph{arXiv preprint arXiv:2203.12602}, 2022.

\bibitem[Wu et~al.(2018)Wu, Zaheer, Hu, Manmatha, Smola, and
  Kr{\"a}henb{\"u}hl]{wu2018compressed}
Wu, C.-Y., Zaheer, M., Hu, H., Manmatha, R., Smola, A.~J., and
  Kr{\"a}henb{\"u}hl, P.
\newblock Compressed video action recognition.
\newblock In \emph{Proceedings of the IEEE conference on computer vision and
  pattern recognition}, pp.\  6026--6035, 2018.

\bibitem[Wu et~al.(2019{\natexlab{a}})Wu, Xiong, Jiang, and
  Davis]{wu2019liteeval}
Wu, Z., Xiong, C., Jiang, Y.-G., and Davis, L.~S.
\newblock Liteeval: A coarse-to-fine framework for resource efficient video
  recognition.
\newblock In Wallach, H., Larochelle, H., Beygelzimer, A., d\textquotesingle
  Alch\'{e}-Buc, F., Fox, E., and Garnett, R. (eds.), \emph{Advances in Neural
  Information Processing Systems}, volume~32. Curran Associates, Inc.,
  2019{\natexlab{a}}.
\newblock URL
  \url{https://proceedings.neurips.cc/paper/2019/file/bd853b475d59821e100d3d24303d7747-Paper.pdf}.

\bibitem[Wu et~al.(2019{\natexlab{b}})Wu, Xiong, Ma, Socher, and
  Davis]{wu2019adaframe}
Wu, Z., Xiong, C., Ma, C.-Y., Socher, R., and Davis, L.~S.
\newblock Adaframe: Adaptive frame selection for fast video recognition.
\newblock In \emph{Proceedings of the IEEE/CVF Conference on Computer Vision
  and Pattern Recognition}, pp.\  1278--1287, 2019{\natexlab{b}}.

\bibitem[Xu et~al.(2017)Xu, Zhao, Xiao, Wu, Zhang, He, and Zhuang]{xu2017video}
Xu, D., Zhao, Z., Xiao, J., Wu, F., Zhang, H., He, X., and Zhuang, Y.
\newblock Video question answering via gradually refined attention over
  appearance and motion.
\newblock In \emph{Proceedings of the 25th ACM international conference on
  Multimedia}, pp.\  1645--1653, 2017.

\bibitem[Xu et~al.(2021)Xu, Ghosh, Huang, Arora, Aminzadeh, Feichtenhofer,
  Metze, and Zettlemoyer]{xu2021vlm}
Xu, H., Ghosh, G., Huang, P.-Y., Arora, P., Aminzadeh, M., Feichtenhofer, C.,
  Metze, F., and Zettlemoyer, L.
\newblock Vlm: Task-agnostic video-language model pre-training for video
  understanding.
\newblock In \emph{Findings of the Association for Computational Linguistics:
  ACL-IJCNLP 2021}, pp.\  4227--4239, 2021.

\bibitem[Yeung et~al.(2016)Yeung, Russakovsky, Mori, and Fei-Fei]{yeung2016end}
Yeung, S., Russakovsky, O., Mori, G., and Fei-Fei, L.
\newblock End-to-end learning of action detection from frame glimpses in
  videos.
\newblock In \emph{Proceedings of the IEEE conference on computer vision and
  pattern recognition}, pp.\  2678--2687, 2016.

\bibitem[Zhang et~al.(2016)Zhang, Wang, Wang, Qiao, and Wang]{zhang2016real}
Zhang, B., Wang, L., Wang, Z., Qiao, Y., and Wang, H.
\newblock Real-time action recognition with enhanced motion vector cnns.
\newblock In \emph{Proceedings of the IEEE conference on computer vision and
  pattern recognition}, pp.\  2718--2726, 2016.

\bibitem[Zhang et~al.(2020{\natexlab{a}})Zhang, Sun, Jing, and
  Zhou]{zhang2020span}
Zhang, H., Sun, A., Jing, W., and Zhou, J.~T.
\newblock Span-based localizing network for natural language video
  localization.
\newblock In \emph{Proceedings of the 58th Annual Meeting of the Association
  for Computational Linguistics}, pp.\  6543--6554, 2020{\natexlab{a}}.

\bibitem[Zhang et~al.(2021)Zhang, Sun, Jing, Zhen, Zhou, and
  Goh]{zhang2021natural}
Zhang, H., Sun, A., Jing, W., Zhen, L., Zhou, J.~T., and Goh, R. S.~M.
\newblock Natural language video localization: A revisit in span-based question
  answering framework.
\newblock \emph{IEEE transactions on pattern analysis and machine
  intelligence}, 2021.

\bibitem[Zhang et~al.(2020{\natexlab{b}})Zhang, Peng, Fu, and
  Luo]{zhang2020learning}
Zhang, S., Peng, H., Fu, J., and Luo, J.
\newblock Learning 2d temporal adjacent networks for moment localization with
  natural language.
\newblock In \emph{Proceedings of the AAAI Conference on Artificial
  Intelligence}, volume~34, pp.\  12870--12877, 2020{\natexlab{b}}.

\bibitem[Zhou \& Berg(2015)Zhou and Berg]{zhou2015temporal}
Zhou, Y. and Berg, T.~L.
\newblock Temporal perception and prediction in ego-centric video.
\newblock In \emph{Proceedings of the IEEE International Conference on Computer
  Vision}, pp.\  4498--4506, 2015.

\bibitem[Zhu et~al.(2020)Zhu, Tran, Sevilla-Lara, Yang, Feiszli, and
  Wang]{zhu2020faster}
Zhu, L., Tran, D., Sevilla-Lara, L., Yang, Y., Feiszli, M., and Wang, H.
\newblock Faster recurrent networks for efficient video classification.
\newblock In \emph{Proceedings of the AAAI Conference on Artificial
  Intelligence}, volume~34, pp.\  13098--13105, 2020.

\end{thebibliography}
\bibliographystyle{icml2023}

\newpage
\appendix
\section*{Appendix: Table of contents}
\parbox{0.5cm}{\textbf{\ref{suppsec:em_task_archs}.}} Description of EM methods \\
\parbox{0.5cm}{\textbf{\ref{suppsec:em_losses}.}} SpotEM distillation losses \\
\parbox{0.5cm}{\textbf{\ref{suppsec:gumbel_max}.}} Gumbel-softmax for clip sampling \\
\parbox{0.5cm}{\textbf{\ref{suppsec:implementation}.}} Implementation details \\
\parbox{0.5cm}{\textbf{\ref{suppsec:flops_calc}.}} Calculating computational cost for SpotEM \\
\parbox{0.5cm}{\textbf{\ref{suppsec:full_results}.}} Complete results for EgoVLP, ReLER \\
\parbox{0.5cm}{\textbf{\ref{suppsec:rio_ablation}.}} Ablation study of RIO features \\
\parbox{0.5cm}{\textbf{\ref{suppsec:video_duration}.}} NLQ performance vs. video duration \\
\parbox{0.5cm}{\textbf{\ref{suppsec:spotem_behavior}.}} Clip-selection behaviors of SpotEM \\
\parbox{0.5cm}{\textbf{\ref{suppsec:qualitative}.}} Qualitative analysis of SpotEM \\
\parbox{0.5cm}{\textbf{\ref{suppsec:tacos_results}.}} Benchmarking SpotEM on \tacos{} dataset \\

\section{Description of EM methods}
\label{suppsec:em_task_archs}

\begin{figure}[t]
    \centering
    \includegraphics[width=\textwidth,clip,trim={0 8cm 10cm 0}]{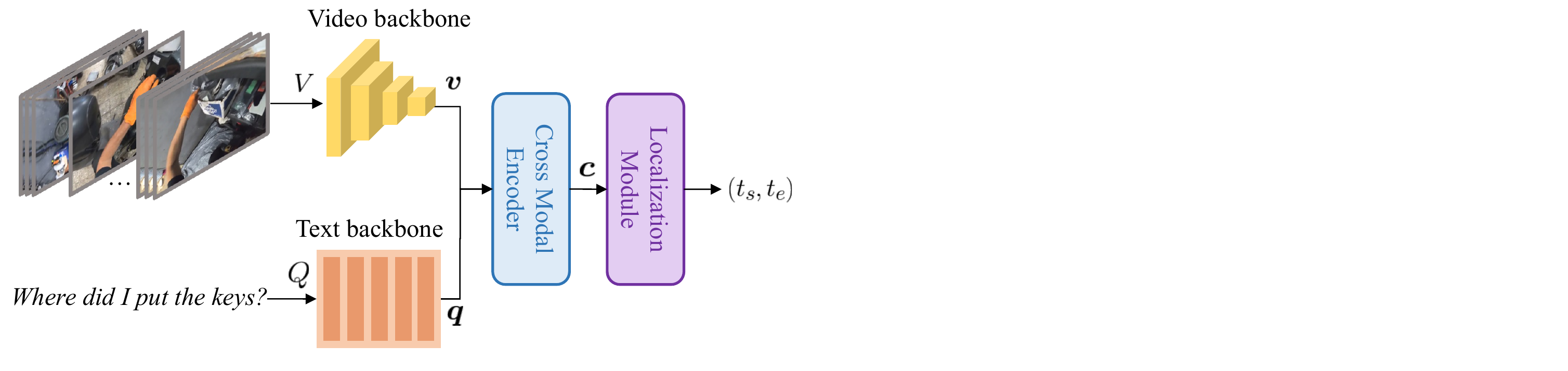}
    \caption{\textbf{Episodic Memory architecture:} We depict how an abstracted episodic memory model works for the natural language queries task. Given video $V$ and text query $Q$, it extracts corresponding features $\vfeat$ and $\qfeat$ using pretrained backbone models. A cross-modal encoder is used to jointly reason across the two modalities and obtain a cross-modal embedding $\bm{c}$. The localization model predicts the location of the response conditioned on the cross-modal embedding.}
    \label{suppfig:em_arch}
\end{figure}

We experiment with three different EM methods in the main paper~\cite{liu2022reler,lin2022egocentric,chen2022internvideo}. \cref{suppfig:em_arch} depicts the working of an abstracted episodic memory architecture that encapsulates these three methods. We now provide more details about the individual methods. 

\textbf{InternVideo}~\cite{chen2022internvideo} proposes a video foundation model that learns a single video representation to achieve state-of-the-art on several tasks including EM. It pretrains two VideoMAE video backbones~\cite{tong2022videomae} to predict verbs and nouns associated with Ego4D clips, respectively. Overall, it combines the EgoVLP TimeSformer, VideoMAE-verb, and VideoMAE-noun backbones to obtain visual representations. It extracts query-text features using the DistillBERT backbone pretrained by~\citet{lin2022egocentric}. It uses the VSLNet architecture for temporal localization~\cite{zhang2020span}, where the cross-modal encoder consists of a transformer encoder to independently encode video and query features, and a context-query attention module to obtain cross-modal embeddings~\cite{seo2016bidirectional}. A span-based localizer is used to localize the response. It consists of a highlight predictor that predicts the probability of each clip overlapping with the response, and uses normalized probabilities to re-weight the video features for localization. A localizer the predicts the start-end probabilities for the resonse conditioned on the output of the highlight module. Please see~\cref{suppsec:em_losses} for more details. This method won the ECCV 2022 EM challenge. 

\textbf{ReLER}~\cite{liu2022reler} uses a modified version of VSLNet-L~\cite{zhang2021natural} and proposes video-level data augmentation techniques for NLQ. It uses a TimeSformer backbone pretrained from~\cite{lin2022egocentric} and a pretrained CLIP image encoder~\cite{radford2021learning} as video encoders. The text backbone is a pretrained CLIP text encoder. It adapts the VSLNet-L architecture for temporal localization~\cite{zhang2021natural}.  The cross-modal encoder consists of a multi-scale transformer encoder to encode video clips, a transformer encoder for the query, and a cross-modal attention mechanism to obtain cross-modal embeddings. A span-based localizer is used as the localization module (similar to VSLNet). Please see~\cite{liu2022reler} for more details. This method was the winning entry in the CVPR 2022 EM challenge. 

\textbf{EgoVLP}~\cite{lin2022egocentric} performs large-scale egocentric video-language pretraining on paired (video clip, narrations text) from Ego4D. This method placed second in the CVPR 2022 EM challenge. It pretrains a TimeSformer~\cite{bertasius2021space} video backbone and a DistillBERT~\cite{sanh2019distilbert} text backbone. Similar to InternVideo, EgoVLP uses the VSLNet architecture for temporal localization~\cite{zhang2020span}. 

\section{SpotEM distillation losses}
\label{suppsec:em_losses}

We now describe the VSLNet architecture~\cite{zhang2020span}, its EM task losses, and how we apply our distillation losses to this model (refer~\cref{sec:optim}).

\subsection*{VSLNet architecture}

We first overview the architecture of VSLNet. Specifically, we use the variant proposed by~\citet{grauman2022ego4d}, where the recurrent modules in the original model from~\citet{zhang2020span} are replaced by transformer modules. Let $\vfeat$ and $\qfeat$ denote the video and query features obtained from the respective backbones (see~\cref{suppfig:em_arch}). A cross-modal encoder is used to perform cross-modal reasoning:
\begin{equation*}
    \cfeat = \textrm{CrossModalEncoder}(\vfeat, \qfeat) \in \mathbb{R}^{L\times D_h}.
\end{equation*}
This consists of a transformer encoder module that independently updates the video features $v$ and query features $q$ by performing self-attention. It then uses the context-query attention mechanism to enhance the video features using information from the query features~\cite{zhang2020span,seo2016bidirectional}.  

VSLNet then introduces the query-guided highlighting (QGH) module, which is a 1D convolutional layer that predicts the probability that a clip lies within a temporal neighborhood of the response: 
\begin{equation}
    \label{suppeq:highlight}
    \hat{\mathcal{S}}_h = \sigma(\textrm{Conv1D}(\cfeat)) \in \mathbb{R}^{L\times 1},
\end{equation}
where $\sigma$ is the sigmoid activation function. These probabilities are used to re-weight the cross-modal features $\cfeat$: 
\begin{equation*}
    \cfeat_h = \hat{\mathcal{S}}_h \cdot \cfeat \in \mathbb{R}^{L \times D_h}.
\end{equation*}
Finally, VSLNet uses a conditioned span predictor to infer the probabilities of each feature location being the start and end points of the response window:
\begin{equation}
    \label{suppeq:span}
    \hat{p}_s, \hat{p}_e = \textrm{ConditionalSpanPredictor}(\cfeat_h),
\end{equation}
where $\hat{p}_s, \hat{p}_e \in \mathbb{R}^{L\times 1}$ are log-probabilities per feature location and ``ConditionalSpanPredictor" consists of a transformer encoder for performing self-attention and an MLP to predict the log-probabilities.  Next, we describe the loss functions proposed in VSLNet for the EM task.

\subsection*{EM task losses}
VSLNet trains the model using a span loss and a QGH loss. The span loss is used to supervise the start/end probability predictions from~\cref{suppeq:span}:
\begin{equation*}
   \mathcal{L}_{\textrm{span}} = \frac{1}{2}\big[f_\textrm{CE}(\hat{p}_s, p^{*}_s) + f_\textrm{CE}(\hat{p}_e, p^{*}_e)\big],
\end{equation*}
where $p^{*}_s, p^{*}_e$ are the ground-truth labels for the response start and end times, and $f_\textrm{CE}$ is the cross-entropy loss function. The QGH loss is used to supervise the query-guided highlighting prediction from~\cref{suppeq:highlight}:
\begin{equation*}
   \mathcal{L}_{\textrm{QGH}} = f_\textrm{CE}(\hat{S}_h, S^{*}_h ),
\end{equation*}
where $S^{*}_h$ is the ground-truth highlight score that covers an extended temporal window around the ground-truth response $\mathcal{R}$. Please see the work from~\citet{zhang2020span} for more details. The overall EM loss combines the span and QGH losses: $\mathcal{L}_\textrm{EM} = \mathcal{L}_{\textrm{span}} + \mathcal{L}_{\textrm{QGH}}$.

\subsection*{Distillation losses} In~\cref{sec:optim}, we introduced our distillation losses to break the negative feedback loop when jointly training MemorySpotter and the EM model. Specifically, we trained an expert EM model that performs the task with all the video and semantic index features without any clip sampling (i.e., no MemorySpotter). We then derived supervision from the expert EM model in the form of distillation losses to train our student model. We proposed the feature distillation loss $L_\textrm{FD}$ to train the student model to match the expert's cross-modal features (see~\cref{eqn:fd_loss}). We also proposed the prediction distillation loss $L_\textrm{PD}$ to encourage the student's localization predictions to match those of the expert. 

We now provide more details about $L_\textrm{PD}$ in the context of VSLNet. Let us denote the highlight scores, start and end log-probabilities predicted by the expert as $S_h^\textrm{expert}$, $p_s^\textrm{expert}$ and $p_e^\textrm{expert}$, respectively~\footnote{The~~$\hat{}$~~symbol is ignored for brevity.} (refer to Equations~\ref{suppeq:highlight} and~\ref{suppeq:span}). Let $S_h^\textrm{student},~p_s^\textrm{student},~p_e^\textrm{student}$ be the corresponding predictions from the student model. Then, the prediction distillation loss is defined as follows:
\begin{align}
    \label{suppeq:pd_loss}
    \begin{split}
    L_\textrm{PD} & = \lambda_\textrm{PD}^h D_\textrm{KL}\Big(S_h^\textrm{student}~||~S_h^\textrm{expert}\Big) \\
    & +~\lambda_\textrm{PD}^l D_\textrm{KL}\Big(\phi(p_s^\textrm{student})~||~\phi(p_s^\textrm{expert})\Big) \\
    & +~\lambda_\textrm{PD}^l D_\textrm{KL}\Big(\phi(p_e^\textrm{student})~||~\phi(p_e^\textrm{expert})\Big),
    \end{split}
\end{align}
where $\phi$ is the softmax activation to convert $p_s$ and $p_e$ into probabilities, $\lambda_\textrm{PD}^{h}$ is the loss scaling for highlight scores, $\lambda_\textrm{PD}^{l}$ is the loss scaling for localization predictions, and $D_\textrm{KL}$ is the KL divergence between two probability distributions.

\section{Gumbel-softmax for clip sampling}
\label{suppsec:gumbel_max}

As discussed in~\cref{sec:optim}, MemorySpotter predicts binary selection values in~\cref{eqn:binarypred} and is therefore non-differentiable for gradient-based optimization. Prior work has tackled this issue using black-box policy optimization, i.e., reinforcement learning~\cite{yeung2016end,wu2019adaframe} or the gumbel-softmax trick which uses a softmax relaxation of argmax sampling during training for differentiability~\cite{hazan2012partition,maddison2017concrete,jang2017categorical,wu2019liteeval}. In this work, we adopt the gumbel-softmax formulation since RL has known issues such as high-variance and instability in training, and requires carefully designed reward functions. 

We now formally describe how we use gubmel-softmax trick for training our clip-selection model by adapting notations from~\citet{jang2017categorical}. Let $b_l$ be a binary variable specifying whether clip $l$ should be sampled given a video-query pair $(V, Q)$ (refer to~\cref{eqn:binarypred}).  Let $\pi^l$ be the probability of selecting clip $l$ predicted by the SelectionPolicy. We can then use the gumbel-softmax trick to approximately draw samples $b_l$ from the 2-class categorical distribution with probabilities $[\pi_0, \pi_1]$, where $\pi_0 = \pi^l$ is the probability of sampling clip $l$ and $\pi_1 = 1 - \pi^l$ is the probability of ignoring clip $l$. The gumbel-softmax distribution is obtained as follows: 
\begin{equation}
    \label{suppeq:gs_approx}
    y_i = \frac{\textrm{exp}\Big((\textrm{log} (\pi_i) + g_i) / \tau\Big)}{\sum_{j\in\{0, 1\}} \textrm{exp}\Big((\textrm{log} (\pi_j) + g_j) / \tau\Big)},
\end{equation}
where $i \in [0, 1]$ and $\tau$ is the softmax temperature. We then use the ``straight-through gumbel-softmax estimator" from~\cite{jang2017categorical} for training.

During the forward pass, discrete samples $b_l$ are drawn through argmax sampling from the gumbel-softmax distribution in~\cref{suppeq:gs_approx}.
\begin{equation}
\begin{split}
    \bar{b}_l &= \textrm{one\_hot}\big(\textrm{argmax} (y_0, y_1)\big) \\
    b_l &= \bar{b}_l[0] \\
\end{split}
\end{equation}
This is done for each clip $l \in [1, \cdots, L]$ to obtain the clip selections $\bfeatb$ from~\cref{eqn:binarypred}. 

During the backward pass, the gumbel-softmax approximation in~\cref{suppeq:gs_approx} is used to compute the gradients. Specifically, we use the selection loss from~\cref{eqn:selection_loss} to encourage the model to select a specified budget of clips and penalize under-/over-sampling. The underlying autograd algorithm (pytorch, in our case) maintains the logits from the forward pass (under the hood) for calculating gradients during the backward pass.

\section{Implementation details}
\label{suppsec:implementation}

We implement all experiments in PyTorch. We modify the implementations of base EM methods to incorporate our SpotEM model and loss functions. For InternVideo and EgoVLP, we use the official NLQ repository released by~\citet{grauman2022ego4d}. For ReLER, we use the code released by~\citet{liu2022reler}. Unlike the former methods, ReLER performs video-level augmentation, where clips from other videos are randomly concatenated on either sides of a given video. We found it unsuitable to train clip sampling methods like SpotEM and LiteEval using this augmentation since the clip selection budget used in the loss function from~\cref{eqn:selection_loss} is inconsistent between training and inference. For example, if the model is trained to select $10\%$ of clips during training, it may pick more than $10\%$ of clips during inference since there are several irrelevant clips appended to the video during training. Therefore, we pre-train the EM modules for SpotEM and LiteEval using uniform clip sampling + video-level augmentation first, and then learn the clip sampling policies in a second stage of training without video-level augmentation. This was not needed for OCSampler since it deterministically picks the top-k clips during inference (instead of deciding the number of clips to select based on a loss function). We provide the hyperparameters for training SpotEM in~\cref{supptab:hyperparameters}.

\begin{table}[t]
\centering
\resizebox{0.48\textwidth}{!}{
\begin{tabular}{@{}rcc@{}}
\toprule
                                   & InternVideo / EgoVLP       &       ReLER             \\ \midrule
Optimizer                          & AdamW                      &       AdamW             \\
Encoder hidden size                & 128                        &        256              \\
\# recursive steps ($N$)           & 4                          &         4               \\
\# training epochs                 & 200                        &        200              \\
Batch size                         & 128                        &        128              \\
Learning rate scheduler            & Linear Warmup              &         -               \\
Initial learning rate              & 0.001                      &       0.0004            \\
$\lambda_\textrm{SEL}$             & [300.0, 1000.0]            &  [30.0, 100.0, 300.0]   \\
$\lambda_\textrm{FD}$              & 1.0                        &        3.0              \\
$\lambda_\textrm{PD}$              & 1.0                        &        1.0              \\
$\lambda_\textrm{PD}^l$            & 1.0                        &        1.0              \\
$\lambda_\textrm{PD}^h$            & [10.0, 30.0]               &       10.0              \\
\bottomrule
\end{tabular}
}
\vspace*{-0.07in}
\caption{\textbf{Hyperparameters for training SpotEM.} $\lambda_\textrm{SEL}, \lambda_\textrm{FD}, \lambda_\textrm{PD}$ are loss scaling hyperparameters from~\cref{eqn:loss_distill}. $\lambda_\textrm{PD}^l, \lambda_\textrm{PD}^h$ are loss scaling hyperparameters from~\cref{suppeq:pd_loss}. For hyperparameters with multiple values, we perform a grid-search over the specified values and pick the best model based on validation performance.}
\label{supptab:hyperparameters}
\end{table}

\section{Calculating computational cost for SpotEM}
\label{suppsec:flops_calc}

In~\cref{tab:internvideo_results,tab:egovlp_results,tab:reler_results} and~\cref{fig:plot_results}, we reported the computational cost in terms of averaged TFLOPs per (video, query) pairs. Here, we present a complete breakdown of the compute costs during inference for SpotEM.
The overall cost $\ccost$ for a video-query pair $(\video, \query)$ is calculated as follows:
\begin{equation}
   \ccost =  \ccost_v \times N + \ccost_s \times L + \ccost_{cs} + \ccost_{em},
\end{equation}
where $N$ is the number of clips selected by the SpotEM, and $L$ is the total number of video clips. The individual cost terms are described as follows:

\parbox{0.6cm}{$\ccost_v$} = extract video features per clip (Equation~\ref{eqn:videobackbone}) \\
\parbox{0.6cm}{$\ccost_s$} = extract semantic-index per image (Equation~\ref{eqn:indexbackbone}) \\
\parbox{0.6cm}{$\ccost_{cs}$} = recursive clip selection for $(\video, \query)$ (Equations~\ref{eqn:crossmodal_cs}, \ref{eqn:binarypred}) \\
\parbox{0.6cm}{$\ccost_{em}$} = EM inference for $(\video, \query)$ (Equations~\ref{eqn:textbackbone}, \ref{eqn:crossmodal}, \ref{eqn:localize})

InternVideo and EgoVLP use the VSLNet model with maximum clip length $L = 128$. ReLER uses the VSLNet-L model with maximum clip length $L = 600$. The cost values for SpotEM when integrated with different base EM methods are shown in~\cref{supptab:ccost}.

\begin{table}[t]
\centering
\resizebox{0.40\textwidth}{!}{%
\begin{tabular}{@{}cccc@{}}
\toprule
               & \multicolumn{3}{c}{Computational cost (in GFLOPs)}       \\
Base EM method & \thead{$\ccost_v$\\(per clip)} & \thead{$\ccost_s$\\(per image)} & \thead{$\ccost_{cs} + \ccost_{em}$\\(per $(\video, \query)$)} \\ \midrule
EgoVLP         & 185.8      & 2.3        & 7.27                        \\
ReLER          & 220.9      & 2.3        & 215.5                       \\
InternVideo    & 2090.8     & 2.3        & 7.27                        \\ \bottomrule
\end{tabular}%
}
\caption{Computational cost breakdown for SpotEM}
\label{supptab:ccost}
\end{table}
\section{Complete results for EgoVLP, ReLER}
\label{suppsec:full_results}

Analogous to~\cref{tab:internvideo_results} from the main paper, we present results on EgoVLP and ReLER methods in~\cref{tab:egovlp_results,tab:reler_results}. We observe trends that are similar to InternVideo, with the exception of SpotEM w/o distill underperforming with ImageNet features for ReLER.

\begin{table}[t]
\resizebox{0.48\textwidth}{!}{
\begin{tabular}{@{}ccccccc@{}}
\toprule
 Row    &Clip selection method   & $\eta$     & Sem. index     &     MR@1    &     MR@5    & TFLOPs    \\ \midrule
        &   ZeroClips            & 100        &    ImageNet    &     3.62    &     8.19    &   0.1     \\ \midrule
   1    &   Random               & 90         &    ImageNet    &     3.79    &     8.23    &   2.4    \\
   2    &   Uniform              & 90         &    ImageNet    &     4.29    &     9.02    &   2.4     \\
   3    &   LiteEval~\cite{wu2019liteeval} &90&    ImageNet    &     5.08    &    10.13    &   2.2     \\
   4    &   OCSampler~\cite{lin2022ocsampler}&90&  ImageNet    &     5.17    &    10.62    &   2.4     \\
   5    &   SpotEM w/o distill (ours) & 91         &    ImageNet    &     6.00    &    12.04    &   2.1     \\ 
   6    &   SpotEM w/o distill (ours)  & 90   &      RIO       &     7.08    &    13.73    &   2.6     \\ 
   7    &   SpotEM (ours)        & 90         &      RIO       & \tb{8.37}   &\tb{15.26}   &   2.6     \\ \midrule
   1    &   Random               & 75         &    ImageNet    &     5.15    &    10.75    &   6.0     \\
   2    &   Uniform              & 75         &    ImageNet    &     5.32    &    10.23    &   6.0     \\
   3    &   LiteEval~\cite{wu2019liteeval} & 75 &  ImageNet    &     5.40    &    11.56    &   5.9     \\
   4    &   OCSampler~\cite{lin2022ocsampler}&75&  ImageNet    &     6.60    &    12.17    &   6.0     \\
   5    &   SpotEM w/o dilstill (ours) & 77         &    ImageNet    &     7.34    &    13.53    &   5.5     \\
   6    &   SpotEM w/o distill (ours) & 76    &      RIO       &     8.15    &    15.59    &   5.9     \\ 
   7    &   SpotEM (ours)        & 75         &      RIO       & \tb{9.17}   &\tb{16.44}   &   6.1     \\ \midrule
   1    &   Random               & 50         &    ImageNet    &     6.36    &    12.07    &  11.9     \\
   2    &   Uniform              & 50         &    ImageNet    &     6.97    &    13.67    &  11.9     \\
   3    &   LiteEval~\cite{wu2019liteeval} & 50 &  ImageNet    &     6.74    &    13.27    &  11.9     \\
   4    &   OCSampler~\cite{lin2022ocsampler}&50&  ImageNet    &     6.74    &    12.68    &  11.9     \\
   5    &   SpotEM w/o distill (ours) & 51         &    ImageNet    &     8.36    &    15.68    &  11.8     \\
   6    &   SpotEM w/o distill (ours) & 52    &      RIO       &     8.44    &    15.89    &  11.8     \\ 
   7    &   SpotEM (ours)        & 51         &      RIO       & \tb{9.29}   &\tb{17.08}   &  11.9     \\ \midrule
 \rowcolor[HTML]{C0C0C0}
        & AllClips~\cite{lin2022egocentric} & 0  &    -     &     9.53    &    17.50    &  23.7     \\ \bottomrule
\end{tabular}
}
\vspace*{-0.1in}
\caption{Comparing efficient clip selection methods for the EgoVLP NLQ method~\cite{lin2022egocentric} on the Ego4D NLQ benchmark.}
\label{tab:egovlp_results}
\end{table}

\begin{table}[t]
\resizebox{0.48\textwidth}{!}{
\begin{tabular}{@{}ccccccc@{}}
\toprule
 Row    &   Clip selection method& $\eta$     & LW feats.      &     MR@1    &     MR@5    & TFLOPs    \\ \midrule
        &   ZeroClips            & 100        &    ImageNet    &      5.66   &      7.52   &    0.3    \\ \midrule
   1    &   Random               & 90         &    ImageNet    &      6.05   &      7.89   &    3.1    \\
   2    &   Uniform              & 90         &    ImageNet    &      6.23   &      7.95   &    3.1    \\
   3    &   LiteEval~\cite{wu2019liteeval} &90&    ImageNet    &      7.16   &      9.10   &    3.1    \\
   4    &   OCSampler~\cite{lin2022ocsampler}&90&  ImageNet    &      8.64   &     11.14   &    3.1    \\
   5    & SpotEM w/o distill (ours) & 90         &    ImageNet    &      8.19   &     10.60   &    2.9    \\ 
   6    & SpotEM w/o distill (ours) & 90      &     ORInt      &      9.58   &     12.05   &    3.3    \\ 
   7    &   SpotEM (ours)        & 90         &     ORInt      &  \tb{9.79}  & \tb{12.16}  &     3.3   \\ \midrule
   1    &   Random               & 75         &    ImageNet    &      7.31   &      9.36   &    7.3    \\
   2    &   Uniform              & 75         &    ImageNet    &      8.19   &     10.59   &    7.3    \\
   3    &   LiteEval~\cite{wu2019liteeval} & 75 &  ImageNet    &      8.81   &     11.50   &    7.2    \\
   4    &   OCSampler~\cite{lin2022ocsampler}&75&  ImageNet    &      9.20   &     11.65   &    7.3    \\
   5    &SpotEM w/o distill (ours)& 77         &    ImageNet    &      9.59   &     12.32   &    6.9    \\
   6    &SpotEM w/o distill (ours)& 75        &     ORInt      &     10.33   &     12.70   &    7.5    \\ 
   7    &   SpotEM (ours)        & 75         &     ORInt      & \tb{10.85}  & \tb{13.89}  &    7.5    \\ \midrule
   1    &   Random               & 50         &    ImageNet    &      9.33   &     11.65   &   14.4    \\
   2    &   Uniform              & 50         &    ImageNet    &      9.46   &     12.46   &   14.4    \\
   3    &   LiteEval~\cite{wu2019liteeval} & 50 &  ImageNet    &     10.37   &     13.06   &   14.3    \\
   4    &   OCSampler~\cite{lin2022ocsampler}&50&  ImageNet    &      9.87   &     12.41   &   14.4    \\
   5    &SpotEM w/o distill (ours)& 51         &    ImageNet    &     10.18   &     12.67   &   14.1    \\
   6    &SpotEM w/o distill (ours)&50         &     ORInt      &     10.58   &     13.49   &   14.5    \\ 
   7    &   SpotEM (ours)        & 50         &     ORInt      & \tb{11.19}  & \tb{14.16}  &   14.6    \\ \midrule
 \rowcolor[HTML]{C0C0C0}
 &AllClips~\cite{liu2022reler}& 0          &         -      &     11.50   &     14.65   &   28.4    \\ \bottomrule
\end{tabular}
}
\vspace*{-0.1in}
\caption{Comparing efficient clip selection methods for the ReLER NLQ method~\cite{liu2022reler} on the Ego4D NLQ benchmark.} 
\label{tab:reler_results}
\end{table}

\section{Ablation study of RIO features}
\label{suppsec:rio_ablation}

In~\cref{tab:ablation_feats}, we studied the effect of RIO features on the overall task performance. We observed that the each of the features were important for obtaining our best performance, especially at higher efficiency levels. We now perform a more detailed study about how these features impact different query types. The NLQ dataset from~\citet{grauman2022ego4d} was constructed based on 13 templates spanning questions about objects, places and people. We evaluate the performance of our SpotEM method and the impact of removing different features on each template in~\cref{supptab:rio_ablation}. Since the features are most impactful at higher efficiency levels (as noted in~\cref{tab:ablation_feats}), we directly evaluate with $\eta=90$. We only select 10/13 templates that have atleast 100 queries in the validation set. Each of the RIO features has a varying impact on the templates. Removing room features affects 3/10 templates where the query object has strong scene association (e.g., ovens are in kitchens). Removing interaction queries affects 7/10 templates which require reasoning about object-interactions (e.g., \emph{``how many drawers did I open?"}). Finally, removing object features affects 2/10 of queries that require object-oriented reasoning (e.g., \emph{``how many funnels are on the shelf?"}). 

\begin{table*}[t]
\centering
\resizebox{\textwidth}{!}{
\begin{tabular}{@{}ccc|P{2.0cm}P{0.85cm}P{1.4cm}P{1.4cm}P{2.2cm}P{1.2cm}P{1.0cm}P{0.75cm}|P{1.4cm}|P{2.3cm}@{}} & & & \multicolumn{8}{c|}{Object queries} & \multicolumn{1}{c|}{Place queries} & \multicolumn{1}{c}{People queries} \\ \cmidrule{4-13}
R & I & O & {\small Where is X before/after Y?} & {\small Where is X?} & {\small What did I put in X?} & {\small How many X's?} & {\small In what location did I see X?} & {\small What X did I Y?} & {\small What X is Y?} & {\small State?} & {\small Where did I put X} & {\small Who did I interact with during Y?} \\ \midrule
 \yes & \yes & \yes &     16.55 &     9.30 &     18.89 &     23.68 &     8.18 &     17.72 &     11.52 &     16.87 &     11.09 &     13.04 \\
      & \yes & \yes &     14.77 &    10.39 &     16.94 &\tbr{21.05}&     7.43 &\tbr{13.91}& \tbr{8.98}&     19.63 &     11.27 &     13.04 \\
 \yes &      & \yes &\tbr{11.88}&     8.21 &\tbr{13.02}&\tbr{15.79}&\tbr{5.76}&\tbr{9.77} &\tbr{9.18} &     18.40 &     10.64 & \tbr{9.24}\\
 \yes & \yes &      &     15.79 &     8.70 &     16.94 &\tbr{19.47}&     7.62 &\tbr{11.42}&     11.33 &     18.10 &     9.66  &     13.04 \\
\bottomrule
\end{tabular}
}
\vspace*{-0.05in}
\caption{\textbf{Impact of semantic index on different types of queries}: We split the performance of different feature types reported in~\cref{tab:ablation_feats} across query types (shown in row 2). We report the mean recall@5 at the highest efficiency level $\eta = 90$, since the features have the biggest impact at this level. Columns 1-3 indicate the presence of room (R), interaction (I), and object (O) features. Row 3 is our proposed method that uses all features. For rows 4-6 are ablations where one of R/I/O features are removed. We highlight the cases where performance drops by more than 2 points on mean recall (in \tbr{red}).}
\label{supptab:rio_ablation}
\end{table*}

We further study the role of RIO features for efficiently performing EM. Specifically, we are interested in knowing if the features only help SpotEM select clips intelligently, or do they help improve the base EM model's performance as well. To study this, we trained each EM method — InternVideo, ReLER, EgoVLP — by appending the RIO features to the original clip features, and used all clips (i.e., no clip sampling). The results are shown in~\cref{supptab:role-rio}. Across methods, we find that RIO features are not helpful when simply concatenated to the original clip features. Their primary utility is to provide cues for intelligent clip selection to responding to a query.

\begin{table}[t]
\centering
\resizebox{0.3\textwidth}{!}{%
\begin{tabular}{@{}lcc@{}}
\toprule
\multicolumn{1}{c}{Method} &    MR@1  &     MR@5  \\ \midrule
EgoVLP                     &\tb{9.54} & \tb{17.50}\\
EgoVLP + RIO               &    9.33  &     16.45 \\ \midrule
ReLER                      &\tb{11.50}& \tb{14.65}\\
ReLER + RIO                &    10.68 &     13.10 \\ \midrule
InternVideo                &\tb{11.45}& \tb{20.56}\\
InternVideo + RIO          &    10.91 &     18.94 \\ \bottomrule
\end{tabular}%
}
\vspace*{-0.05in}
\caption{\textbf{Role of RIO features for EM:} RIO features are not beneficial when simply concatenated with the base EM model's clip features. Their primary role is to help SpotEM intelligently select clips relevant to the query.}
\label{supptab:role-rio}
\end{table}

\section{NLQ performance vs. video duration}
\label{suppsec:video_duration}

We now study the performance of clip selection methods as a function of the video duration. Longer videos are more challenging since the search space grows significantly and more computational cost may be required for inference. We group the NLQ validation video clips into four buckets: 0-5 mins, 5-10 mikns, 10-15 mins, and 15-20 mins. The statistics of the number of queries and clips in each bucket are shown in~\cref{supptab:video_stats}. Majority of the clips are 5-10 mins long. To get reliable statistics for evaluation (i.e., at least 100 queries), we only evaluate on clips in the 5-10 mins and 15-20 mins buckets. We compare the ``mean R@1" performance on 5-10 mins clips and 15-20 mins clips for the following methods: AllClips (the upper bound baseline), LiteEval, OCSampler, SpotEM. The base EM method is InternVideo. The results are in~\cref{supptab:video_duration_results}. 

We make a few observations. The performance of the AllClips “upper bound” is significantly worse on 15-20 mins clips when compared to 5-10 mins clips, pointing to the difficulty of the task. When sampling only 25\% of the clips, SpotEM recovers $\sim85\%$ of AllClips’ performance for 15 - 20 mins clips. While this reduces from $\sim96\%$ for 5 - 10 mins clips, it is still a relatively high number. Thus, SpotEM continues to work well for longer videos, but the overall performance is ultimately bottlenecked by the underlying EM method. Finally, SpotEM outperforms the learned baselines LiteEval and OCSampler on all cases, confirming its advantages over the baselines.

\begin{table}[t]
\centering
\resizebox{0.48\textwidth}{!}{%
\begin{tabular}{@{}rcccc@{}}
\toprule
\multicolumn{1}{c}{} & 0-5 mins & 5-10 mins & 10-15 mins & 15-20 mins \\ \midrule
\# video clips       & 13       & 289       & 0          & 18         \\
\# queries           & 43       & 3142      & 0          & 344        \\ \bottomrule
\end{tabular}%
}
\vspace*{-0.05in}
\caption{Distribution of queries and video clips as a function of video duration.}
\label{supptab:video_stats}
\end{table}

\begin{table}[t]
\centering
\resizebox{0.5\textwidth}{!}{%
\begin{tabular}{@{}lcccccccc@{}}
\toprule
\multicolumn{1}{c}{}           & \multicolumn{4}{c}{5-10 mins clips} & \multicolumn{4}{c}{15 - 20 mins clips} \\
\multicolumn{1}{r}{Efficiency} & 0       & 50      & 75     & 90     & 0        & 50      & 75      & 90      \\ \cmidrule(lr){1-1}\cmidrule(lr){2-5}\cmidrule(lr){6-9}
AllClips                       & 12.22   & -       & -      & -      & 4.80     & -       & -       & -       \\
LiteEval                       & -       & 9.34    & 7.72   & 6.29   & -        & 2.03    & 1.31    & 1.02    \\
OCSampler                      & -       & 9.79    & 7.72   & 6.02   & -        & 2.62    & 3.34    & 2.47    \\
SpotEM                         & -       &\tb{12.27}&\tb{11.73}&\tb{10.20}& -   &\tb{4.65}&\tb{4.07}&\tb{3.20}\\ \midrule
\multicolumn{1}{r}{\% results} & -       & 100.4   & 95.9   & 83.4   & -        & 96.8    & 84.7    & 66.6    \\ \bottomrule
\end{tabular}%
}
\vspace*{-0.10in}
\caption{\textbf{NLQ accuracy vs.~video duration.} The last row shows the \% of AllClips’ results achieved by SpotEM.}
\label{supptab:video_duration_results}
\end{table}

\section{Clip-selection behaviors of SpotEM}
\label{suppsec:spotem_behavior}

As we motivate in~\cref{sec:intro}, the key idea behind SpotEM is that not all parts of the video are useful for a given query and there are high-level visual semantics that could steer our attention towards where to look (refer~\cref{sec:intro}). Based on this intuition, we decomposed the EM task into two steps: (1) identify query-relevant clips using a high-level preview — the role of SpotEM, and (2) solve the EM task using the selected clips — the role of the base EM model. Importantly, SpotEM is \emph{not} trying to answer the query. Instead, it aims to narrow the search to some relevant clips (and reject several irrelevant clips) and enable the EM model to perform the task using a lower computational budget.

Another reasonable alternative is to directly train an EM model to predict the ground-truth response window using only the semantic index, and use these predicted response windows to select clips for the heavier video feature extraction. Based on this, we created the \emph{direct-supervision} clip selector  (as opposed to SpotEM, where we indirectly supervise clip selection based on the EM performance using the selected clips). It works as follows: (1) train a ZeroClips EM model to predict the GT response using RIO features, (2) given an NLQ query during inference, use the ZeroClips model to infer the top-k responses, and (3) select clips that overlap with the top-k responses for expensive feature computation.  We then train new EM modules (i.e., CrossModalEncoder and LocalizationModule) to perform NLQ using the clips selected by direct-supervision (i.e., RIO features for all clips + video features for selected clips).

The results are in~\cref{supptab:direct-supervision}. When compared to SpotEM, direct-supervision performs poorly. The problem with direct-supervision is that the semantic indexes we use (RIO, in this case) are low resolution in terms of the semantics encoded and may not be enough to infer the GT response accurately. Given a clip with 16 frames (at 30 FPS), the semantic index only encodes a single image within that clip using an efficient image backbone (refer to~\cref{sec:eem_desc}). Therefore, methods that directly predict the GT response using the semantic index perform poorly (i.e., direct-supervision), which eventually translates to poor clip selection. This suggests that selecting query-relevant clips (and more importantly, rejecting irrelevant clips) is critical for efficiently performing EM.

\begin{table}[t]
\centering
\resizebox{0.44\textwidth}{!}{%
\begin{tabular}{@{}cccc@{}}
\toprule
Method                        & Efficiency & MR@1      & MR@5    \\ \midrule
Direct-supervision (top $k$=1) & 90         & 5.27      & 9.50       \\
SpotEM                        & 90         & \tb{7.48} & \tb{14.82} \\
Direct-supervision (top $k$=2)   & 75         & 5.48      & 10.24      \\
SpotEM                        & 75         & \tb{9.92} & \tb{17.27} \\
Direct-supervision (top $k$=5)   & 50         & 7.64      & 13.23      \\
SpotEM                        & 50         & \tb{9.84} & \tb{18.70} \\ \bottomrule
\end{tabular}%
}
\vspace*{-0.05in}
\caption{EM accuracy comparison between direct-supervision and SpotEM. \textbf{Note:} We train both methods without distillation losses for apples-to-apples comparison. For direct-supervision, we train models for $k=1,2,3,4,5$ and select the largest $k$ that satisfies the given efficiency budget.}
\label{supptab:direct-supervision}
\end{table}

To understand the clip-selection behaviors resulting from indirectly supervising the clip selection in SpotEM, we empirically study the relationship between the clips selected by SpotEM and the ground-truth response. For SpotEM trained at 90\% efficiency, we divide the validation queries into cases where the model gets the prediction right (i.e., MR@5 = 1) and the model gets the prediction wrong (i.e., MR@5 $<$ 1). We then measured two statistics for each case.

\textbf{(1) mean IoU} measures the intersection over union between the clips selected by SpotEM and clips belonging to the GT response, averaged over all queries. \\
\textbf{(2) mean nonzero intersection} measures the percentage of queries where SpotEM selects at least one clip belonging to the GT response.

Results are shown in~\cref{supptab:relation_clip_gt}. We make two observations. The mean IoU is low for both correct and wrong cases, i.e., SpotEM does not limit itself to selecting clips within the ground-truth response. The mean nonzero intersection is high for correct cases, i.e., 90\% of the time, SpotEM picks \emph{at least one clip} within the GT response when its prediction is right. These results confirm our intuition that SpotEM does not try to answer the query (since the mean IoU is low). Instead, it rejects irrelevant clips and selects relevant clips (including clips within the GT response).

\begin{table}[t]
\centering
\resizebox{0.45\textwidth}{!}{%
\begin{tabular}{@{}ccc@{}}
\toprule
    SpotEM prediction      & mean IoU $(\times 100)$ & \thead{mean non-zero \\ intersection} \\ \midrule
~right {\small(~~717/3529 queries~)} & 13.96                   & 90.10                      \\
wrong {\small(~2812/3529 queries~)}  & 3.12                    & 32.79                      \\ \bottomrule
\end{tabular}%
}
\vspace*{-0.05in}
\caption{Studying the clip-selection behaviors of SpotEM}
\label{supptab:relation_clip_gt}
\end{table}

\section{Qualitative analysis of SpotEM}
\label{suppsec:qualitative}

Analogous to~\cref{fig:qualitative_2} in the main paper, we visualize five success and two failure cases of \modelname{} in the form of videos.\footnote{Qualitative visualizations are available here: {\scriptsize\url{https://utexas.box.com/s/p8iheclayaoth2ey95m8o0w9m97snjmb}}} In each video, we visualize the clips selected by \modelname{} at each step, highlight their relevance to the query (if any), and provide a textual description of SpotEM's behavior. Finally, we visualize the predicted and ground-truth responses. Additionally, we describe why SpotEM performs poorly in failure cases. In success cases, SpotEM is able to identify query-relevant clips and use them to respond to the query accurately. In failure cases, SpotEM tends to confuse one object for another (e.g., dressing vs. salt container) or confuse object colors (e.g., brown box vs. blue box). These failures could be attributed to the similarity of these objects in the RIO feature space.

\section{Benchmarking SpotEM on \tacos{} dataset}
\label{suppsec:tacos_results}

Ego4D, to the best of our knowledge, is the first dataset to introduce the EM task with unique properties of long-form egocentric videos and natural-language query annotations. To test our approach on another dataset, we identified a third-person video dataset that may support the natural-language grounding task with long video and short responses. The \tacos{} dataset contains long third-person kitchen videos (~5 mins on average) with relatively short natural language moments (~5 secs)~\cite{rohrbach2014coherent}. Since \tacos{} has third-person videos, we use clip features from a SlowFast model pre-trained on Kinetics 400~\cite{feichtenhofer2019slowfast}. Our base method for \tacos{} NLG is a VSLNet model with SlowFast features. We compare SpotEM with all baselines across different efficiency levels in~\cref{supptab:tacos_results}. The trends echo results from~\cref{sec:experiments} on Ego4D NLQ. There are two key differences. LiteEval performs better than OCSampler (row 3 vs. 4), and adding RIO features is not helpful (row 5 vs. 6), likely due to the ego-exo domain shift (RIO features are trained on egocentric images). Overall, SpotEM outperforms the baselines (row 5) and adding distillation losses improves performance by a good margin (row 6 vs. 7). These results confirm the benefits of SpotEM for natural-language grounding on long exocentric videos.

\begin{table}[t]
\resizebox{0.48\textwidth}{!}{
\begin{tabular}{@{}cccccc@{}}
\toprule
 Row    &  Clip selection method              & $\eta$     & Sem. index     &     MR@1    &     MR@5    \\ \midrule
        &   ZeroClips                         & 100        &    ImageNet    &     7.52    &    14.03    \\ \midrule
   1    &   Random                            & 90         &    ImageNet    &     8.09    &    14.42    \\
   2    &   Uniform                           & 90         &    ImageNet    &     8.07    &    13.79    \\
   3    &   LiteEval~\cite{wu2019liteeval}    &90          &    ImageNet    &    12.18    &    18.10    \\
   4    &   OCSampler~\cite{lin2022ocsampler} &90          &    ImageNet    &    10.54    &    17.46    \\
   5    &   \modelname{} w/o distill          & 91         &    ImageNet    &    13.34    &    19.82    \\ 
   6    &   \modelname{} w/o distill          & 90         &      RIO       &    13.66    &    19.95    \\ 
   7    &   \modelname{} (ours)               & 90         &      RIO       &\tb{15.08}   &\tb{21.57}   \\ \midrule
   1    &   Random                            & 75         &    ImageNet    &     9.31    &    17.03    \\
   2    &   Uniform                           & 75         &    ImageNet    &    11.00    &    18.68    \\
   3    &   LiteEval~\cite{wu2019liteeval}    & 75         &    ImageNet    &    14.66    &    21.64    \\
   4    &   OCSampler~\cite{lin2022ocsampler} &75          &    ImageNet    &    12.40    &    18.90    \\
   5    &   \modelname{} w/o distill          & 76         &    ImageNet    &    15.25    &    23.60    \\
   6    &   \modelname{} w/o distill          & 75         &      RIO       &    15.34    &    23.61    \\ 
   7    &   \modelname{} (ours)               & 76         &      RIO       &\tb{17.12}   &\tb{24.98}   \\ \midrule
   1    &   Random                            & 50         &    ImageNet    &    12.96    &    21.00    \\
   2    &   Uniform                           & 50         &    ImageNet    &    14.66    &    23.97    \\
   3    &   LiteEval~\cite{wu2019liteeval}    & 50         &    ImageNet    &    15.56    &    23.44    \\
   4    &   OCSampler~\cite{lin2022ocsampler} &50          &    ImageNet    &    13.56    &    20.46    \\
   5    &   \modelname{} w/o distill          & 53         &    ImageNet    &    16.44    &    24.81    \\
   6    &   \modelname{} w/o distill          & 52         &      RIO       &    16.71    &    25.07    \\ 
   7    &   \modelname{} (ours)               & 51         &      RIO       &\tb{17.90}   &\tb{26.84}   \\ \midrule
 \rowcolor[HTML]{C0C0C0}
        &   AllClips~\cite{grauman2022ego4d}  & 0         &  -             &    18.27    &    26.54     \\ \bottomrule
\end{tabular}
}
\vspace*{-0.05in}
\caption{Comparing efficient clip selection methods on \tacos{} NLG for the VSLNet baseline with SlowFast features~\cite{feichtenhofer2019slowfast}. 
}
\label{supptab:tacos_results}
\end{table}


\end{document}